\documentclass{article}

\usepackage[nonatbib,final]{nips_2018}

\usepackage[utf8]{inputenc} 
\usepackage[T1]{fontenc}    
\usepackage{hyperref}       
\usepackage{url}            
\usepackage{booktabs}       
\usepackage{amsfonts}       
\usepackage{nicefrac}       
\usepackage{microtype}      
\usepackage[title]{appendix}

\usepackage[cmex10]{amsmath}
\usepackage{amssymb,amsthm}
\usepackage{array}
\usepackage[backend=biber,style=ieee,natbib=false,mincrossrefs=1000]{biblatex}
\usepackage{bm}
\usepackage[font={normalsize}]{caption} 
\usepackage{graphicx}
\usepackage{makecell}
\usepackage{subfig}
\usepackage{tabulary}

\newcommand{\E}{\operatorname{\mathbb E}}

\newcommand{\etal}{et al. }
\newcommand{\ie}{i.e.}
\newcommand{\eg}{e.g.}

\newcommand{\tines}{\!\,\times\!\,}

\newcolumntype{P}[1]{>{\centering\arraybackslash}p{#1}}

\title{Joint Autoregressive and Hierarchical Priors for Learned Image Compression}

\author{
  David Minnen, Johannes Ballé, George Toderici\\
  Google Research\\
  \texttt{\{dminnen, jballe, gtoderici\}@google.com}\\
}

\begin{document}

\maketitle

\begin{abstract}
Recent models for learned image compression are based on autoencoders, learning 
approximately invertible mappings from pixels to a quantized latent
representation. These are combined with an entropy model, a prior 
on the latent representation that can be used with standard arithmetic coding algorithms
to yield a compressed bitstream. Recently, hierarchical entropy models have been
introduced as a way to exploit more structure in the latents than simple fully 
factorized priors, improving compression performance while maintaining end-to-end 
optimization. Inspired by the success of autoregressive priors in probabilistic generative models, 
we examine autoregressive, hierarchical, as well as combined priors as alternatives,
weighing their costs and benefits in the context of image compression. While
it is well known that autoregressive models come with a significant computational penalty,
we find that in terms of compression performance, autoregressive and hierarchical 
priors are complementary and, together, exploit the 
probabilistic structure in the latents better than all previous learned models. 
The combined model yields state-of-the-art rate--distortion performance, providing a 
15.8\% average reduction in file size over the previous state-of-the-art method based 
on deep learning, which corresponds to a 59.8\% size reduction over JPEG, 
more than 35\% reduction compared to WebP and JPEG2000, and bitstreams 8.4\% smaller
than BPG, the current state-of-the-art image codec. To the best of our knowledge, 
our model is the first learning-based method to outperform BPG on both PSNR and MS-SSIM
distortion metrics.

\end{abstract}

\section{Introduction}
Most recent methods for learning-based, lossy image compression adopt an approach based on \textit{transform coding}~\cite{goyal2001}. In this approach, image compression is achieved by first mapping pixel data into a quantized latent representation and then losslessly compressing the latents. Within the deep learning research community, the transforms typically take the form of convolutional neural networks (CNNs), which approximate nonlinear functions with the potential to map pixels into a more compressible latent space than the linear transforms used by traditional image codecs. This nonlinear transform coding method resembles an autoencoder~\cite{rumelhart1986, hinton2006}, which consists of an \emph{encoder} transform between the data (in this case, pixels) and a latent, reduced-dimensionality space, and a \emph{decoder}, an approximate inverse function that maps latents back to pixels. While dimensionality reduction can be seen as a simplistic form of compression, it is not equivalent to it, as the goal of compression is to reduce the entropy of the representation under a prior probability model shared between the sender and the receiver (the entropy model), not only the dimensionality. To improve compression performance, recent methods have given increased focus to this part of the model~\cite{toderici2017cvpr, theis2017iclr, balle2017iclr, li2017importance, johnston2018cvpr, rippel2017icml, minnen2017icip, agustsson2017nips, mentzer2018cvpr, balle2018iclr, minnen2018icip}. Finally, the entropy model is used in conjunction with standard entropy coding algorithms such as arithmetic, range, or Huffman coding~\cite{rissanen1981, martin1979, van-leeuwen1976} to generate a compressed bitstream.

The training goal is to minimize the expected length of the bitstream as well as the expected distortion of the reconstructed image with respect to the original, giving rise to a rate--distortion optimization problem:
\begin{equation}
R + \lambda \cdot D = \underbrace{\E_{\bm x \sim p_{\bm x}}\bigl[-\log_2 p_{\bm{\hat y}}(\lfloor{f(\bm x)}\rceil)\bigr]}_{\text{rate}} + \lambda \cdot \underbrace{\E_{\bm x \sim p_{\bm x}}\bigl[d(\bm x, g(\lfloor{f(\bm x)}\rceil))\bigr]}_{\text{distortion}},
\label{eq:rate--distortion}
\end{equation}
where $\lambda$ is the Lagrange multiplier that determines the desired rate--distortion trade-off, $p_{\bm x}$ is the unknown distribution of natural images, $\left\lfloor{\cdot}\right\rceil$ represents rounding to the nearest integer (quantization), $\bm y = f(\bm x)$ is the encoder, $\bm{\hat y} = \left\lfloor{\bm y}\right\rceil$ are the quantized latents, $p_{\bm{\hat y}}$ is a discrete entropy model, and $\bm{\hat x} = g(\bm{\hat y})$ is the decoder with $\bm{\hat x}$ representing the reconstructed image. The rate term corresponds to the cross entropy between the marginal distribution of the latents and the learned entropy model, which is minimized when the two distributions are identical. The distortion term may correspond to a closed-form likelihood, such as when $d(\bm{x},\bm{\hat x})$ represents mean squared error (MSE), which induces an interpretation of the model as a variational autoencoder~\cite{balle2017iclr}. When optimizing the model for other distortion metrics such as MS-SSIM, it is simply minimized as an energy function.

The models we analyze in this paper build on the work of Ballé~\etal\cite{balle2018iclr}, which uses a noise-based relaxation to be able to apply gradient descent methods to the loss function in Eq.~\eqref{eq:rate--distortion} and introduces a hierarchical prior to improve the entropy model. While most previous research uses a fixed, though potentially complex, entropy model, Ballé~\etal use a Gaussian scale mixture (GSM)~\cite{wainwright1999gsm} where the scale parameters are conditioned on a hyperprior. Their model allows for end-to-end training, which includes joint optimization of a quantized representation of the hyperprior, the conditional entropy model, and the base autoencoder. The key insight is that the compressed hyperprior could be added to the generated bitstream as \textit{side information}, which allows the decoder to use the conditional entropy model. In this way, the entropy model itself is image-dependent and spatially adaptive, which allows for a richer and more accurate model. Ballé~\etal show that standard optimization methods for deep neural networks are sufficient to learn a useful balance between the size of the side information and the savings gained from a more accurate entropy model. The resulting compression model provides state-of-the-art image compression results compared to earlier learning-based methods.

We extend this GSM-based entropy model in two ways: first, by generalizing the hierarchical GSM model to a Gaussian mixture model, and, inspired by recent work on generative models, by adding an autoregressive component. We assess the compression performance of both approaches, including variations in the network architectures, and discuss benefits and potential drawbacks of both extensions. For the results in this paper, we did not make efforts to reduce the capacity (\ie, number of channels, layers) of the artificial neural networks to optimize computational complexity, since we are interested in determining the potential of different forms of priors rather than trading off complexity against performance. Note that increasing capacity alone is not sufficient to obtain arbitrarily good compression performance \cite[appendix 6.3]{balle2018iclr}.
\section{Architecture Details}
\begin{figure}[tb]
  \centering
  \begin{minipage}{0.68\textwidth}
    \centering
    \includegraphics[width=\linewidth]{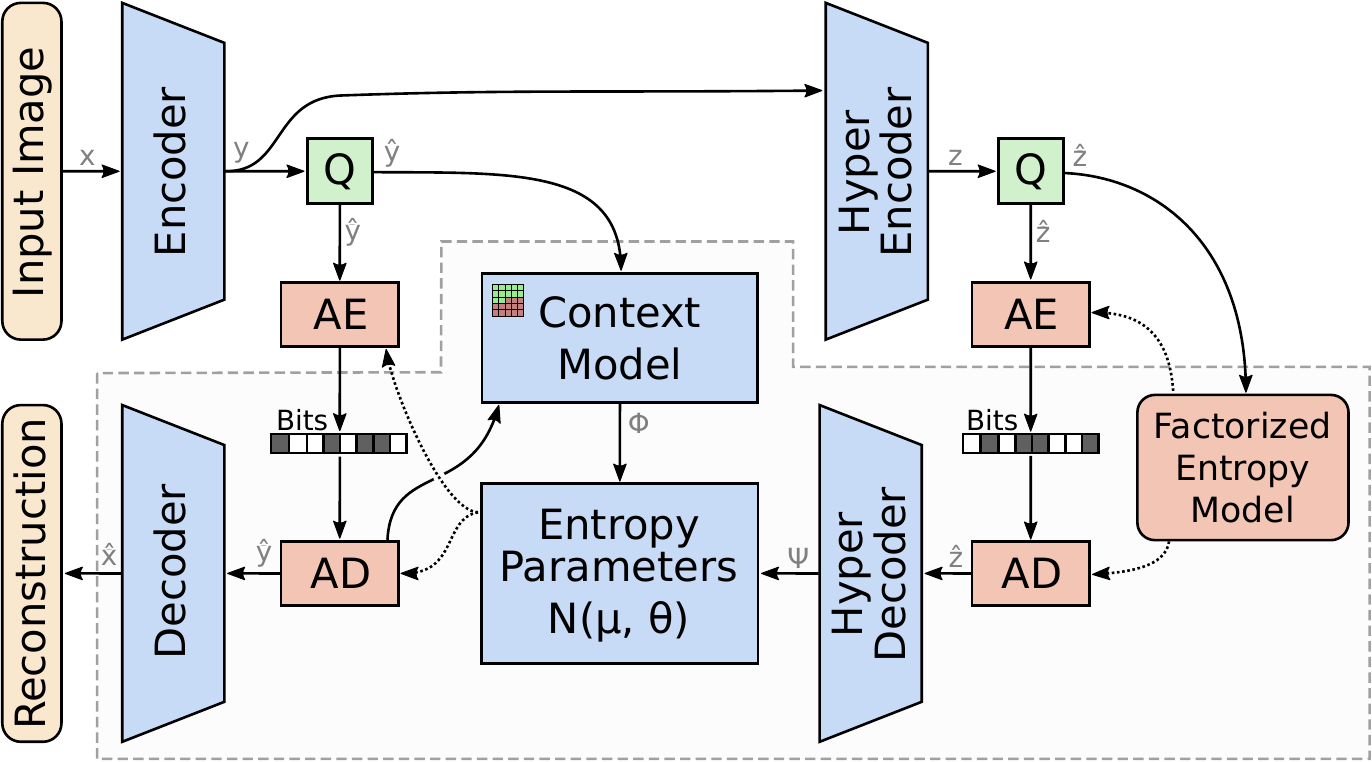}
  \end{minipage}%
  \hfill%
  \begin{minipage}{0.32\textwidth}
    \begin{flushright}
    \scriptsize
    \begin{tabular}[t]{@{\hspace{0.2em}}c@{\hspace{0.7em}}c@{\hspace{0.2em}}}
      \textbf{Component} & \textbf{Symbol} \\ \midrule
      Input Image & $\bm x$ \\
      Encoder & $f(\bm x; \bm \theta_e)$ \\
      Latents & $\bm y$ \\
      Latents (quantized) & $\bm{\hat y}$ \\
      Decoder & $g(\bm{\hat y}; \bm \theta_d)$ \\
      Hyper Encoder & $f_h(\bm y; \bm \theta_{he})$ \\
      Hyper-latents & $\bm z$ \\
      Hyper-latents (quant.) & $\bm{\hat z}$ \\
      Hyper Decoder & $g_h(\bm{\hat z}; \bm \theta_{hd})$ \\
      Context Model & $g_{cm}(\bm y_{<i}; \bm \theta_{cm})$ \\
      Entropy Parameters & $g_{ep}(\cdot; \bm \theta_{ep})$ \\
      Reconstruction & $\bm{\hat x}$
    \end{tabular}%
  \end{flushright}%
  \end{minipage}%
  \caption{Our combined model jointly optimizes an autoregressive component that predicts latents from their causal context (\textit{Context Model}) along with a hyperprior and the underlying autoencoder. Real-valued latent representations are quantized (\textit{Q}) to create latents ($\bm{\hat y}$) and hyper-latents ($\bm{\hat z}$), which are compressed into a bitstream using an arithmetic encoder (\textit{AE}) and decompressed by an arithmetic decoder (\textit{AD}). The highlighted region corresponds to the components that are executed by the receiver to recover an image from a compressed bitstream.}
  \label{fig:architecture}
\end{figure}

Figure~\ref{fig:architecture} provides a high-level overview of our generalized compression model, which contains two main sub-networks\footnote{See Section 4 in the supplemental materials for an in-depth visual comparison between our architecture variants and previous learning-based methods.}.
The first is the core autoencoder, which learns a quantized latent representation of images
(\textit{Encoder} and \textit{Decoder} blocks). The second sub-network is
responsible for learning a probabilistic model over quantized latents used for
entropy coding. It combines the \textit{Context Model}, an autoregressive
model over latents, with the hyper-network (\textit{Hyper Encoder} and
\textit{Hyper Decoder} blocks), which learns to represent information useful
for correcting the context-based predictions. The data from these two sources
is combined by the \textit{Entropy Parameters} network, which generates the
mean and scale parameters for a conditional Gaussian entropy model. 

Once training is complete, a valid compression model must prevent any information from passing between the encoder to the decoder unless that information is available in the compressed file. In Figure~\ref{fig:architecture}, the arithmetic encoding (AE) blocks produce the compressed representation of the symbols coming from the quantizer, which is stored in a file. Therefore at decoding time, any information that depends on the quantized latents may be used by the decoder once it has been decoded. In order for the context model to work, at any point it can only access the latents that have already been decoded. When starting to decode an image, we assume that the previously decoded latents have all been set to zero.

The learning problem is to minimize the expected rate--distortion loss defined in
Eq.~\ref{eq:rate--distortion} over the model parameters. Following the work of
Ballé~\etal\cite{balle2018iclr}, we model each latent, $\hat y_i$, as a Gaussian
convolved with a unit uniform distribution. This ensures a
good match between encoder and decoder distributions of both the quantized latents, and
continuous-valued latents subjected to additive uniform noise during training. While
\cite{balle2018iclr} predicted the scale of each Gaussian conditioned on the
hyperprior, $\bm{\hat z}$, we extend the model by predicting the mean and scale
parameters conditioned on both the hyperprior as well as the causal context of
each latent $\hat y_i$, which we denote $\bm{\hat y}_{<i}$. The predicted Gaussian parameters are functions of the learned parameters of the hyper-decoder, context model, and entropy parameters networks ($\bm \theta_{hd}$, $\bm \theta_{cm}$, and $\bm \theta_{ep}$, respectively):
\begin{multline}
p_{\bm{\hat y}}(\bm{\hat y} \mid \bm{\hat z}, \bm \theta_{hd}, \bm \theta_{cm}, \bm \theta_{ep}) = \prod_i \Bigl( \mathcal N\bigl(\mu_i, \sigma_i^2\bigr) \ast \mathcal U\bigl(-\tfrac 1 2, \tfrac 1 2\bigr) \Bigr)(\hat y_i) \\
\text{with } \mu_i, \sigma_i = g_{ep}(\bm \psi, \bm \phi_i; \bm \theta_{ep}), \bm \psi = g_{h}(\bm{\hat z}; \bm \theta_{hd}), \text{and } \bm \phi_i = g_{cm}(\bm{\hat y}_{<i}; \bm \theta_{cm}).
\label{eq:entropy-model}
\end{multline}

The entropy model for the hyperprior is the same as in~\cite{balle2018iclr},
although we expect the hyper-encoder and hyper-decoder to learn significantly
different functions in our combined model, since they now work in conjunction with an
autoregressive network to predict the entropy model parameters. Since we do
not make any assumptions about the distribution of the hyper-latents, a
non-parametric, fully factorized density model is used. A more powerful
entropy model for the hyper-latents may improve compression rates, e.g., we could
stack multiple instances of our contextual model, but we expect the net effect
to be minimal since $\bm{\hat z}$ comprises only a very small percentage of the
total file size. Because both the compressed latents and the compressed
hyper-latents are part of the generated bitstream, the rate--distortion loss from
Equation~\ref{eq:rate--distortion} must be expanded to include the cost of
transmitting $\bm{\hat z}$. Coupled with a squared error distortion metric, the
full loss function becomes:
\begin{equation}
R + \lambda \cdot D =
\underbrace{\E_{\bm x \sim p_{\bm x}} \bigl[-\log_2 p_{\bm{\hat y}}({\bm{\hat y}})\bigr]}_{\text{rate (latents)}} +
\underbrace{\E_{\bm x \sim p_{\bm x}} \bigl[-\log_2 p_{\bm{\hat z}}({\bm{\hat z}})\bigr]}_{\text{rate (hyper-latents)}}
+ \lambda \cdot
\underbrace{\E_{\bm x \sim p_{\bm x}} \lVert \bm x - \bm{\hat x} \rVert_2^2}_{\text{distortion}}
\label{eq:rd-latents-side}
\end{equation}

\subsection{Layer Details and Constraints}
\label{sec:layers}
Details about the individual network layers in each component of our models are
outlined in Table~\ref{table:layers}. While the internal structure of the
components is fairly unrestricted, \eg, one could exchange the convolutional
layers for residual blocks or dilated convolution without fundamentally changing the model, certain components must be constrained to ensure that availability of the bitstreams alone is sufficient for the receiver to reconstruct the image.

The last layer of the encoder corresponds to the bottleneck of the base autoencoder. Its number of output channels determines the number of elements that must be compressed and stored. Depending on the rate--distortion trade-off, our models learn to ignore certain channels by deterministically generating the same latent value and assigning it a probability of 1, which wastes computation but requires no additional entropy. This modeling flexibility allows us to set the bottleneck larger than necessary, and let the model determine the number of channels that yield the best performance. Similar to reported in other work, we found that too few channels in the bottleneck can impede rate--distortion performance when training models to target higher bit rates, but too having too many does not harm the compression performance.

The final layer of the decoder must have three channels to generate RGB
images, and the final layer of the \textit{Entropy Parameters} sub-network
must have exactly twice as many channels as the bottleneck. This constraint
arises because the \textit{Entropy Parameters} network predicts two values,
the mean and scale of a Gaussian distribution, for each latent. The number of output channels of the \textit{Context Model} and \textit{Hyper Decoder} components are
not constrained, but we also set them to twice the bottleneck size in all of
our experiments.

Although the formal definition of our model allows the autoregressive
component to condition its predictions $\bm \phi_i =
g_{cm}(\bm{\hat y}_{<i}; \bm \theta_{cm})$ on all previous latents, in practice we use a limited context
(5$\times$5 convolution kernels) with masked convolution similar to the approach used by
PixelCNN~\cite{vanDenOord2016pixelcnn}. The \textit{Entropy Parameters}
network is also constrained, since it can not access predictions from the
\textit{Context Model} beyond the current latent element. For simplicity, we use
1$\times$1 convolution in the \textit{Entropy Parameters} network, although
masked convolution is also permissible. Section~\ref{sec:experiments} provides
an empirical evaluation of the model variants we assessed, exploring the effects of different context sizes and more complex
autoregressive networks.

\begin{table}[tb]
  \centering
  \scriptsize
  \setlength\extrarowheight{1pt}
  \setlength\tabcolsep{3pt}
  \begin{tabulary}{\textwidth}{@{}P{21mm}|C|C|P{23mm}|P{23mm}|P{20mm}@{}}
    \thead{Encoder}
    & \thead{Decoder}
    & \thead{Hyper\\Encoder}
    & \thead{Hyper\\Decoder}
    & \thead{Context\\Prediction}
    & \thead{Entropy\\Parameters}
    \\ \midrule
    Conv: 5$\tines$5 c192 s2
    & Deconv: 5$\tines$5 c192 s2
    & Conv: 3$\tines$3 c192 s1
    & Deconv: 5$\tines$5 c192 s2
    & Masked: 5$\tines$5 c384 s1
    & Conv: 1$\tines$1 c640 s1
    \\
    GDN & IGDN & Leaky ReLU & Leaky ReLU & & Leaky ReLU\\
    Conv: 5$\tines$5 c192 s2
    & Deconv: 5$\tines$5 c192 s2
    & Conv: 5$\tines$5 c192 s2
    & Deconv: 5$\tines$5 c288 s2
    &
    & Conv: 1$\tines$1 c512 s1
    \\
    GDN & IGDN & Leaky ReLU & Leaky ReLU & & Leaky ReLU\\
    Conv: 5$\tines$5 c192 s2
    & Deconv: 5$\tines$5 c192 s2
    & Conv: 5$\tines$5 c192 s2
    & Deconv: 3$\tines$3 c384 s1
    &
    & Conv: 1$\tines$1 c384 s1
    \\
    GDN & IGDN & & & &\\
    Conv: 5$\tines$5 c192 s2
    & Deconv: 5$\tines$5 c3 s2
    & & & &
    \\
  \end{tabulary}
  \caption{Each row corresponds to a layer of our generalized model. Convolutional layers
    are specified with the ``Conv'' prefix followed by the kernel size, number
    of channels and downsampling stride (\eg, the first layer of the encoder uses
    5$\times$5 kernels with 192 channels and a stride of two). The ``Deconv''
    prefix corresponds to upsampled convolutions (i.e., in TensorFlow, \texttt{tf.conv2d\_transpose}), while
    ``Masked'' corresponds to masked convolution as
    in~\cite{vanDenOord2016pixelcnn}. GDN stands for generalized divisive
    normalization, and IGDN is inverse GDN~\cite{balle2016gdn}.}
  \label{table:layers}
\end{table}

\section{Experimental Results}
\label{sec:experiments}
\begin{figure}[tb]
  \centering
  \includegraphics[width=0.49\linewidth]{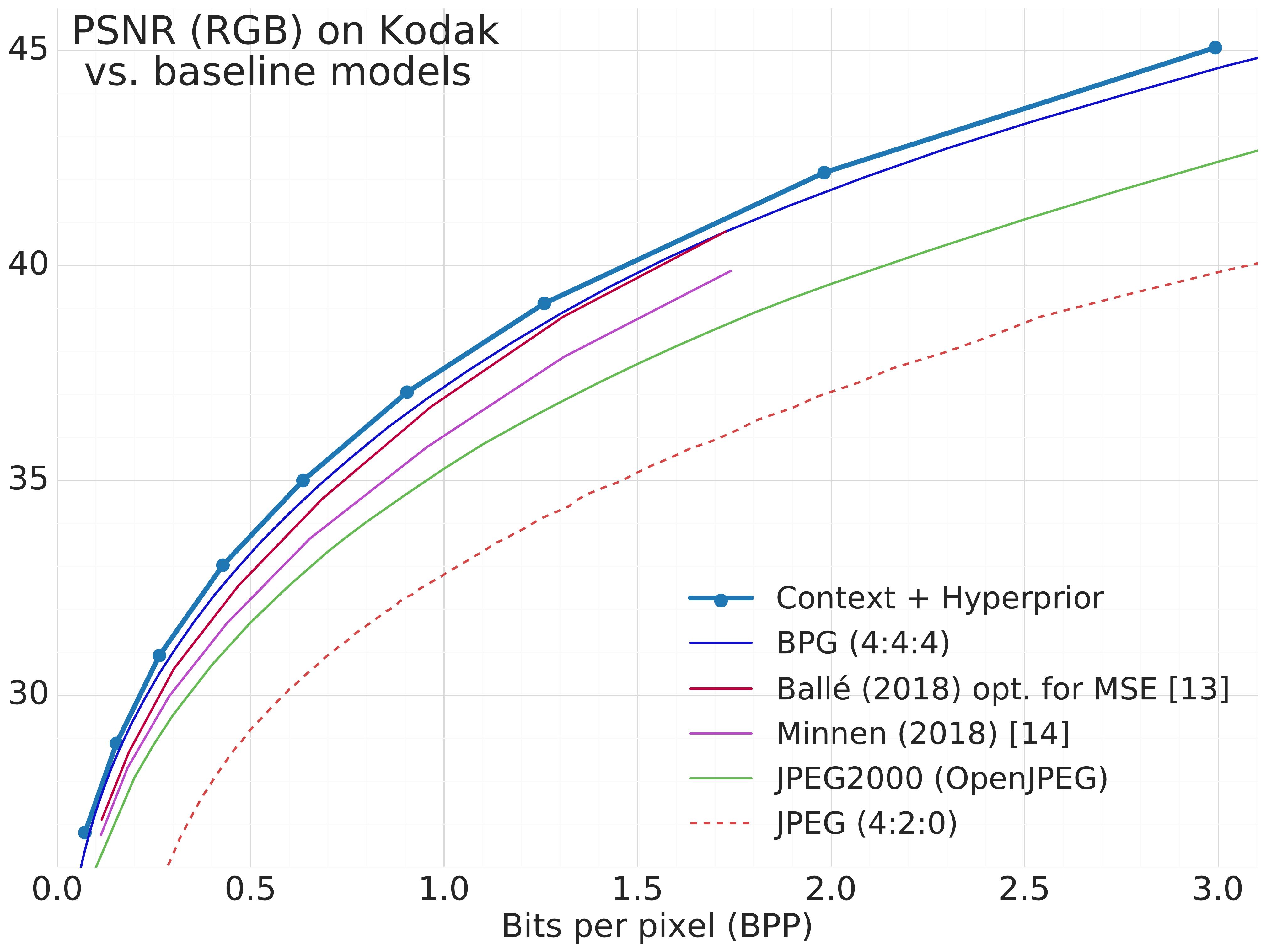}
  \hfill
  \includegraphics[width=0.49\linewidth]{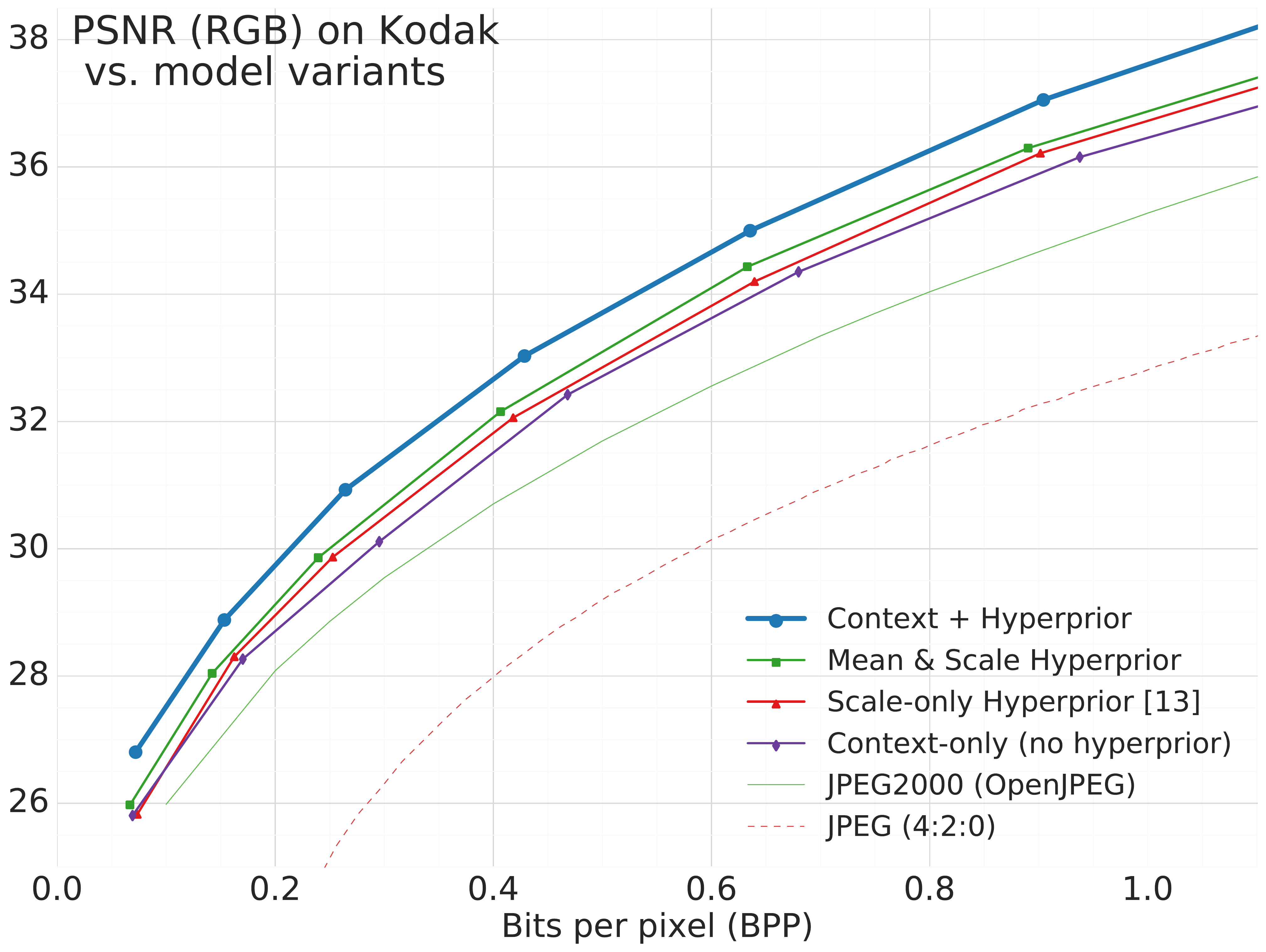}
  \caption{Our combined approach (context + hyperprior) has better rate--distortion performance on the Kodak
    image set as measured by PSNR (RGB) compared to all of the baselines methods (\textit{left}). To our knowledge, this is the first learning-based method to outperform BPG on PSNR. The right graph compares the relative performance of different versions of our method. It shows that using a hyperprior is better than a purely autoregressive (context-only) approach and that combining both (context + hyperprior) yields the best RD performance.}
  \label{fig:rd-kodak-psnr}
\end{figure}

We evaluate our generalized models by calculating the rate--distortion (RD) performance
averaged over the publicly available Kodak image
set~\cite{kodak}\footnote{Please see the supplemental material for additional
evaluation results including full-page RD curves, example images, and results on the
larger Tecnick image set (100 images with resolution 1200$\times$1200).}.
Figure~\ref{fig:rd-kodak-psnr} shows RD curves using peak signal-to-noise
ratio (PSNR) as the image quality metric. While PSNR is known to be a
relatively poor perceptual metric~\cite{tid2013}, it is still a standard
metric used to evaluate image compression algorithms and is the primary metric
used for tuning conventional compression methods. The RD graph on the left of
Figure~\ref{fig:rd-kodak-psnr} compares our combined context + hyperprior model
to existing image codecs (standard codecs and learned models) and
shows that this model outperforms all of the existing methods including
BPG~\cite{bpg}, a state-of-the-art codec based on the intra-frame coding
algorithm from HEVC~\cite{hevc}. To the best of our knowledge, it is
the first learning-based compression model to outperform BPG on PSNR. The
right RD graph compares different versions of our models and shows that the combined model performs the best, while the context-only model performs slightly worse than either hierachical version.

\begin{figure}[tb]
  \centering
  \begin{minipage}{0.5\textwidth}
    \begin{flushleft}
    \includegraphics[width=\linewidth]{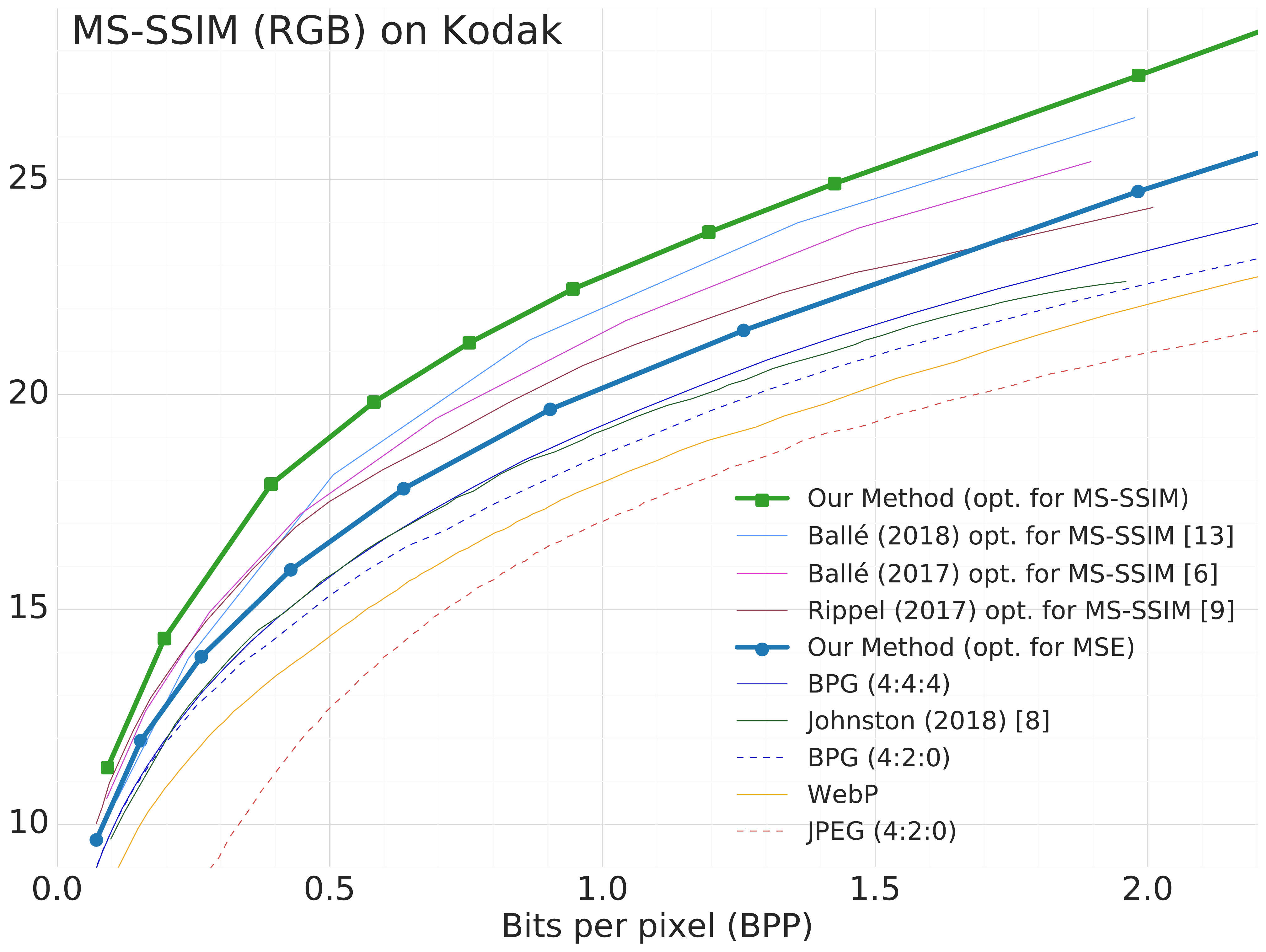}
    \end{flushleft}
    \caption{When evaluated using MS-SSIM (RGB) on Kodak, our combined approach has
      better RD performance than all previous methods when optimized for
      MS-SSIM. When optimized for MSE, our method still provides better
      MS-SSIM scores than all of the standard codecs.}
    \label{fig:kodak-msssim}    
  \end{minipage}%
  \hfill%
  \begin{minipage}{0.47\textwidth}
    \begin{flushright}
    \vspace{-4mm}
    \includegraphics[width=\linewidth]{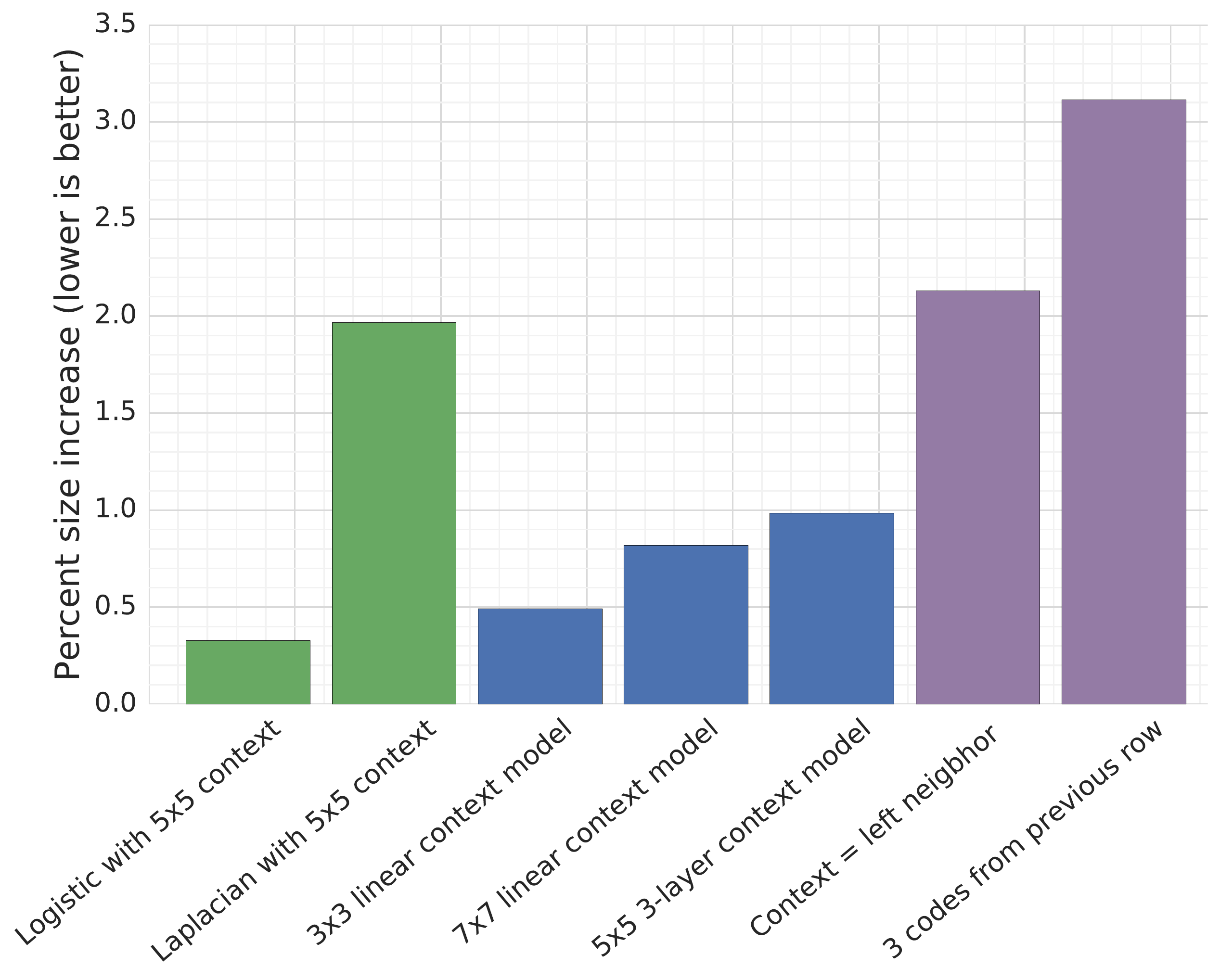}
    \end{flushright}
    \caption{The baseline implementation of our model uses a hyperprior and
      a linear context model with 5$\times$5
      masked convolution. Optimized with $\lambda=0.025$ (bpp $\approx 0.61$ on
      Kodak), the baseline outperforms the other variants we tested (see text
      for details).}
    \label{fig:variants-barchart}
  \end{minipage}%
\end{figure}

Figure~\ref{fig:kodak-msssim} shows RD curves for Kodak using multiscale
structural similarity (MS-SSIM)~\cite{wang2003multiscale} as the image quality
metric. The graph includes two versions of our combined model: one optimized for MSE
and one optimized for MS-SSIM. The latter outperforms all existing methods
including all standard codecs and other learning-based methods that were also
optimized for MS-SSIM (\cite{balle2018iclr, balle2017iclr,
  rippel2017icml}). As expected, when our model is optimized for MSE,
performance according to MS-SSIM falls. Nonetheless, the MS-SSIM scores for
this model still exceed all standard codecs and all learning-based methods
that were not specifically optimized for MS-SSIM.

As outlined in Table~\ref{table:layers}, our baseline architecture for the combined model uses
5$\times$5 masked convolution in a single linear layer for the context model,
and it uses a conditional Gaussian distribution for the entropy
model. Figure~\ref{fig:variants-barchart} compares this baseline
to several variants by showing the relative increase in file
size at a single rate-point. The green bars show that exchanging the Gaussian distribution for a
logistic distribution has almost no effect (the 0.3\% increase is smaller than
the training variance), while switching to a Laplacian distribution decreases
performance more substantially. The blue bars compare different context
configurations. Masked 3$\times$3 and 7$\times$7 convolution both perform
slightly worse, which is surprising since we expected the additional context
provided by the 7$\times$7 kernels to improve prediction accuracy. Similarly,
a 3-layer, nonlinear context model using 5$\times$5 masked convolution also
performed slightly worse than the linear baseline. Finally, the purple bars
show the effect of using a severely restricted context such as only a single
neighbor or three neighbors from the previous row. The primary benefit of
these models is increased parallelization when calculating context-based
predictions since the dependence is reduced from two dimensions down to
one. While both cases show a non-negligible rate increase (2.1\% and 3.1\%,
respectively), the increase may be worthwhile in a practical implementation
where runtime speed is a major concern.

\begin{figure}[b]
  \centering
  \subfloat[Ours (0.2149 bpp)]{\label{lighthouseA}
    \includegraphics[width=0.24\linewidth]{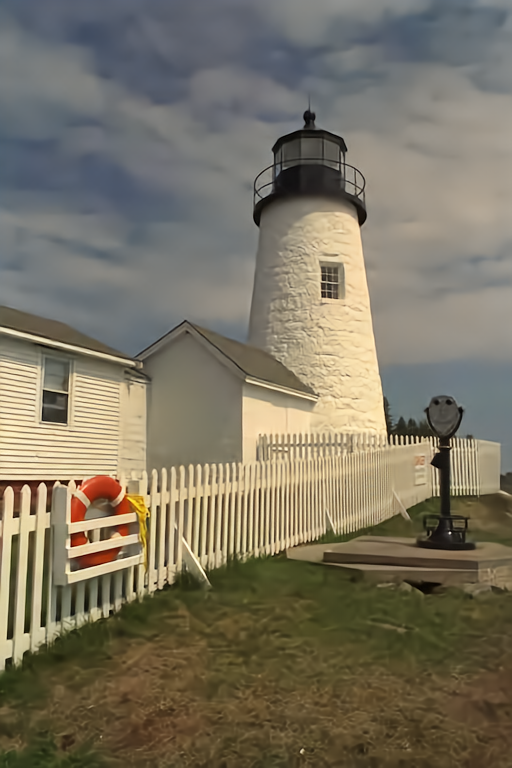}
  }
  \subfloat[Scale-only (0.2205 bpp)]{\label{lighthouseB}
    \includegraphics[width=0.24\linewidth]{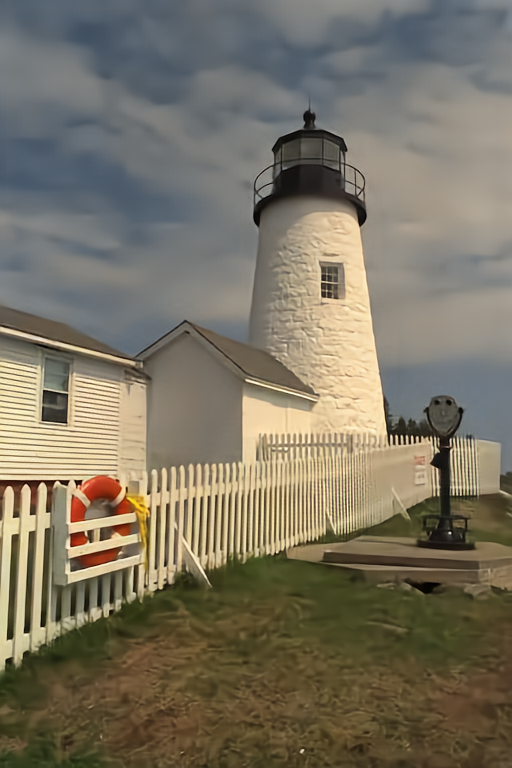}
  }
  \subfloat[BPG (0.2352 bpp)]{\label{lighthouseC}
    \includegraphics[width=0.24\linewidth]{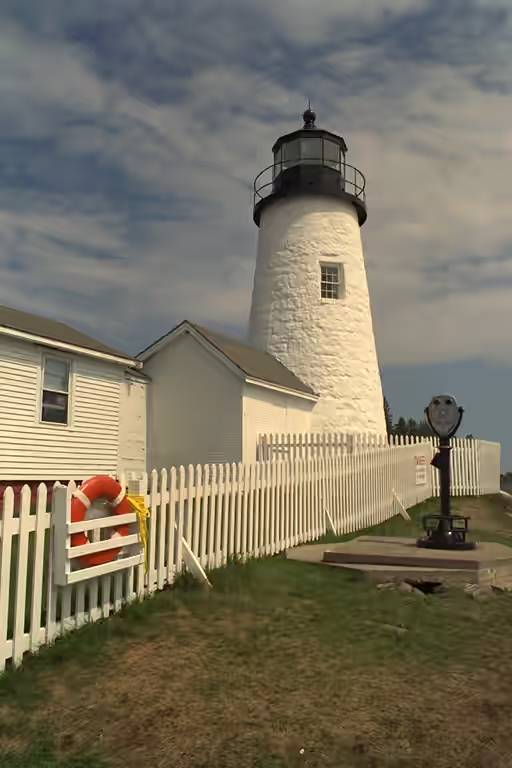}
  }
  \subfloat[JPEG (0.2309 bpp)]{\label{lighthouseD}
    \includegraphics[width=0.24\linewidth]{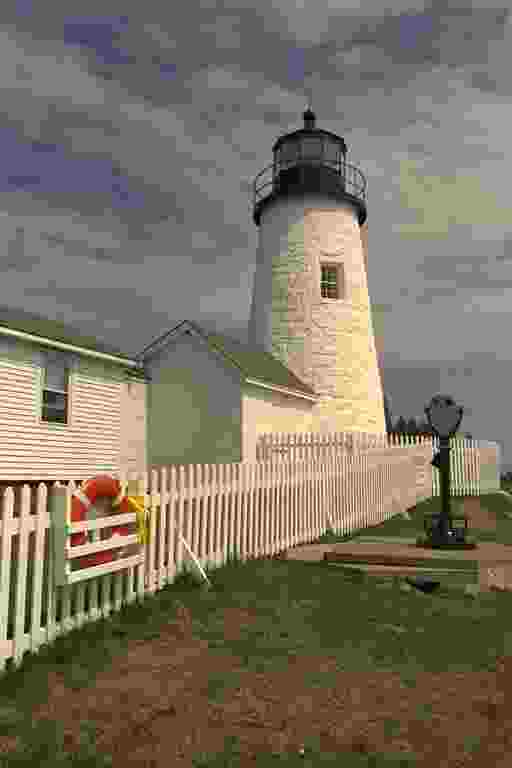}
  }
  \caption{At similar bit rates, our combined method provides the highest visual
    quality. Note the aliasing in the fence in the scale-only version as well
    as a slight global color cast and blurriness in the yellow rope. BPG shows
    more ``classical'' compression artifacts, \eg, ringing around the top of
    the lighthouse and the roof of the middle building. BPG also introduces a
    few geometric artifacts in the sky, though it does preserve extra detail
    in the sky and fence (albeit with 9.4\% more bits that our encoding). JPEG
    shows severe blocking artifacts at this bit rate.}
  \label{fig:lighthouse}
\end{figure}

Finally, Figure~\ref{fig:lighthouse} provides a visual comparison for one of
the Kodak images. Creating accurate comparisons is difficult since most compression methods
do not have the ability to target a precise bit rate. We therefore selected
comparison images with sizes that are as close as possible, but always larger
than our encoding (up to 9.4\% larger in the case of BPG). Nonetheless, our
compression model provides clearly better visual quality compared to the scale
hyperprior baseline~\cite{balle2018iclr} and JPEG. The perceptual quality
relative to BPG is much closer. For example, BPG preserves mode detail in the
sky and parts of the fence, but at the expense of introducing geometric
artifacts in the sky, mild ringing near the building/sky boundaries, and
some boundary artifacts where neighboring blocks have widely different levels of detail
 (\eg, in the grass and lighthouse).

\section{Related Work}
The earliest research that used neural networks to compress images dates back
to the 1980s and relies on an autoencoder with a small bottleneck using either
uniform quantization~\cite{cottrel1989} or vector
quantization~\cite{luttrell1988, watkins1991}. These approaches sought equal
utilization of the codes and thus did not learn an explicit entropy
model. Considerable research followed these initial models, and Jiang provides
a comprehensive survey covering methods published through
the late 1990s~\cite{Jiang1999}.

More recently, image compression with deep neural networks became a popular
research topic starting with the work of Toderici~\etal\cite{ToOMHwViMi16}
who used a recurrent architecture based on LSTMs to learn multi-rate,
progressive models. Their approach was improved by exploring other recurrent
architectures for the autoencoder, training an LSTM-based entropy model, and
adding a post-process that spatially adapts the bit rate based on the
complexity of the local image content~\cite{toderici2017cvpr,
  johnston2018cvpr}. Related research followed a more traditional image coding
approach and explicitly divided images into patches instead of using a fully
convolutional model~\cite{minnen2017icip, baig2017}. Inspired by modern image
codecs and learned inpainting algorithms, these methods trained a neural
network to predict each image patch from its causal context (in the image space,
not the latent space) before encoding the residual. Similarly, most modern image
compression standards use context to predict pixel values as well as using a
context-adaptive entropy model~\cite{jpeg2000, webp, bpg}.

Many learning-based methods take the form of an autoencoder, and multiple
models are trained to target different bit rates instead of training a single
recurrent model~\cite{balle2017iclr, theis2017iclr, mentzer2018cvpr,
  rippel2017icml, li2017importance, agustsson2017nips, minnen2018icip}. Some
use a fully factorized entropy model~\cite{balle2017iclr, theis2017iclr},
while others make use of context in code space to improve compression
rates~\cite{rippel2017icml, li2017importance, mentzer2018cvpr,
  toderici2017cvpr, johnston2018cvpr}. Other methods do not make use of
context via an autoregressive model and instead rely on side information that
is either predicted by a neural network~\cite{balle2018iclr} or composed of
indices into a (shared) dictionary of non-parametric code distributions used
locally by the entropy coder~\cite{minnen2018icip}.

Learned image compression is also related to Bayesian generative models such as
PixelCNN~\cite{vanDenOord2016pixelcnn}, variational
autoencoders~\cite{kingma2014vae}, PixelVAE~\cite{pixelvae},
$\beta$-VAE~\cite{higgins2017betavae}, and VLAE~\cite{chen2017vlae}. In general, Bayesian image models seek to maximize the \emph{evidence}
$\E_{\bm x \sim p_{\bm x}} \log p(\bm x)$, which is generally intractable, and use the joint likelihood, as in Eq.~\eqref{eq:rate--distortion}, as a lower bound, while compression models directly seek to optimize Eq.~\eqref{eq:rate--distortion}. Research on generative models is typically interested in uncovering semantically meaningful, disentangled latent representations. It has been noted that under certain conditions, compression models are formally equivalent to VAEs~\cite{balle2017iclr, theis2017iclr}. $\beta$-VAEs have a particularly strong connection since $\beta$ controls the trade-off between the data log-likelihood (distortion) and prior (rate), as does $\lambda$ in our formulation, which is derived from classical rate--distortion theory. 

A further major difference are the constraints imposed on compression models by the need to quantize and arithmetically encode the latents, which require certain choices regarding the parametric form of the densities and a transition between continous (differential) and discrete (Shannon) entropies. We can draw
strong conceptual parallels between our models and PixelCNN autoencoders~\cite{vanDenOord2016pixelcnn}, and especially PixelVAE~\cite{pixelvae} and VLAE~\cite{chen2017vlae}, when applied to discrete latents. These models are often evaluated by comparing average likelihoods (which correspond to differential entropies), whereas compression models are typically evaluated by comparing several bit rates (corresponding to Shannon entropies) and distortion values across the rate--distortion frontier, making evaluations more complex.

\section{Discussion}
Our approach extends the work of Ballé~\etal\cite{balle2018iclr} in two ways. First, we generalize the GSM model to a conditional Gaussian mixture model (GMM). Supporting this model is simply a matter of generating both a mean and a scale parameter conditioned on the hyperprior. Intuitively, the average likelihood of the observed latents increases when the center of the conditional Gaussian is closer to the true value and a smaller scale is predicted, i.e., more structure can be exploited by modeling conditional means. The core question is whether or not the benefits of this more sophisticated model outweigh the cost of the associated side information. We showed in Figure~\ref{fig:rd-kodak-psnr} (\textit{right}) that a GMM-based entropy model provides a net benefit and outperforms the simpler GSM-based model in terms of rate--distortion performance without increasing the asymptotic complexity of the model.

The second extension that we explore is the idea of combining an autoregressive model with the hyperprior. Intuitively, we can see how these components are complementary in two ways. First, starting from the perspective of the hyperprior, we see that for identical hyper-network architectures, improvements to the entropy model require more side information. The side information increases the total compressed file size, which limits its benefit. In contrast, introducing an autoregressive component into the prior does not incur a potential rate penalty since the predictions are based only on the causal context,~\ie, on latents that have already been decoded. Similarly, from the perspective of the autoregressive model, we expect some amount of uncertainty that can not be eliminated solely from the causal context. The hyperprior, however, can ``look into the future'' since it is part of the compressed bitstream and is fully known by the decoder. The hyperprior can thus learn to store information needed to reduce the uncertainty in the autoregressive model while avoiding information that can be accurately predicted from context.

\begin{figure}[tb]
  \centering
  \includegraphics[width=\textwidth]{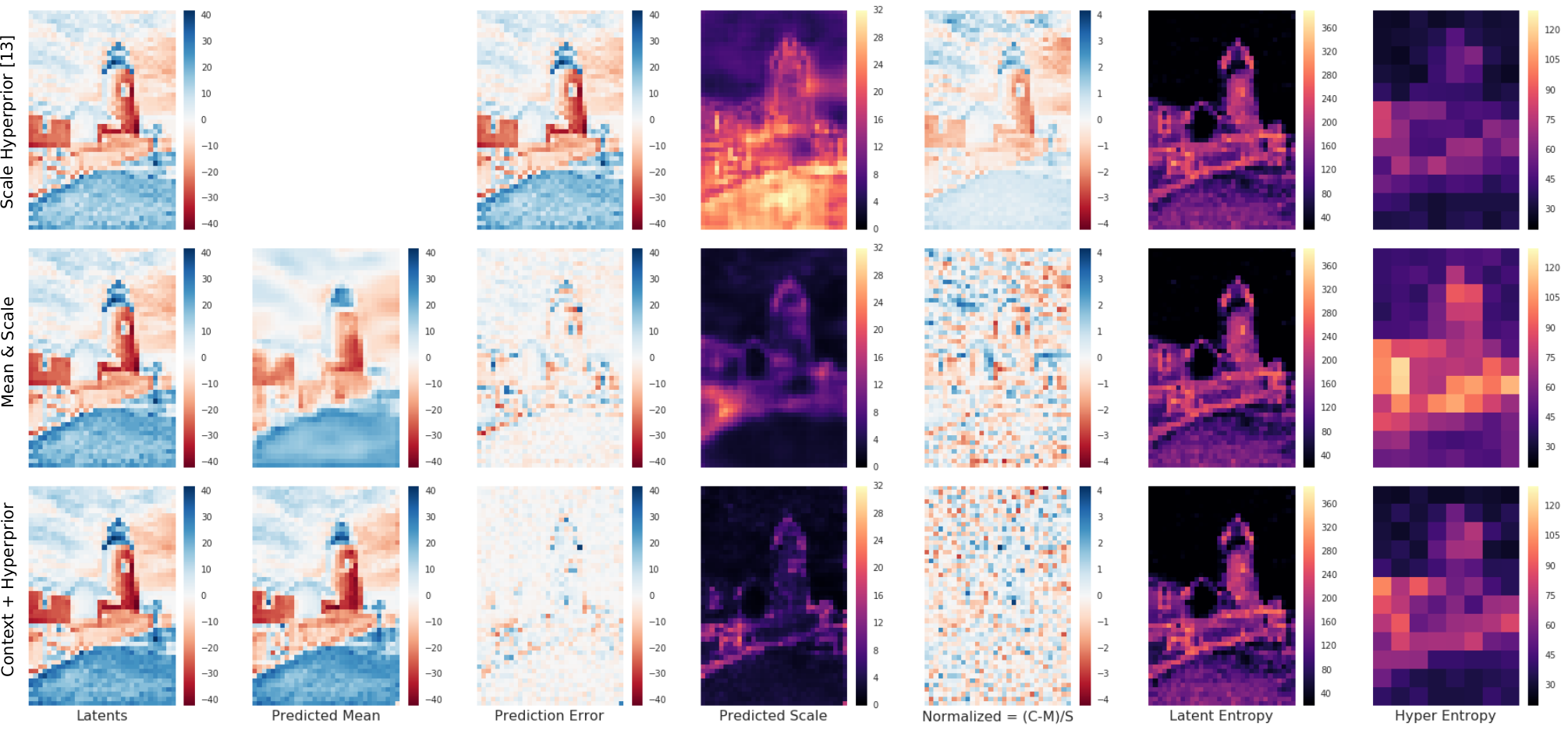}
  \caption{Each row corresponds to a different model variant and shows
    information for the channel with the highest entropy. The visualizations
    show that more powerful models reduce the prediction error, require
    smaller scale parameters, and remove structure from the normalized latents,
    which directly translates into a more accurate entropy model and thus
    higher compression rates.}
  \label{fig:code-viz}
\end{figure}

Figure~\ref{fig:code-viz} visualizes some of the internal mechanisms of our models. We show three of the variants: one Gaussian scale mixture equivalent to~\cite{balle2018iclr}, another strictly hierarchical prior extended to a Gaussian mixture model, and one combined model using an autoregressive component and a hyperprior. After encoding the lighthouse image shown in Figure~\ref{fig:lighthouse}, we extracted the latents for the channel with the highest entropy. These latents are visualized in the first column of Figure~\ref{fig:code-viz}. The second column holds the conditional means and clearly shows the added detail attained with an autoregressive component, which is reminiscent of the observation that VAE-based models tend to produce blurrier images than autoregressive models~\cite{pixelvae}. This improvement leads to a lower prediction error (third column) and smaller predicted scales, i.e. smaller uncertainty (fourth column). Our entropy model assumes that latents are conditionally independent given the hyperprior, which implies that the normalized latents, i.e. values with the predicted mean and scale removed, should be closer to i.i.d. Gaussian noise. The fifth column of Figure~\ref{fig:code-viz} shows that the combined model is closest to this ideal and that the autoregressive model helps significantly (compare row 4 with row 2). Finally, the last two columns show how the entropy is distributed across the image for the latents and hyper-latents.

From a practical standpoint, autoregressive models are less desirable than hierarchical models since they are inherently serial, and therefore can not be sped up using techniques such as parallelization. To report the performance of the compression models which contain an autoregressive component, we refrained from implementing a full decoder for this paper, and instead compare Shannon entropies. We have empirically verified that these measurements are within a fraction of a percent of the size of the bitstream generated by arithmetic coding. 

Probability density distillation has been successfully used to get around the serial nature of autoregressive models for the task of speech synthesis~\cite{oord2017parallel}, but unfortunately the same type of method cannot be applied in the domain of compression due to the coupling between the prior and the arithmetic decoder. To address these computational concerns, we have begun to explore very lightweight context models as described in Section~\ref{sec:experiments} and Figure~\ref{fig:variants-barchart}, and are considering further techniques to reduce the computational requirements of the \textit{Context Model} and \textit{Entropy Parameters} networks, such as engineering a tight integration of the arithmetic decoder with a differentiable autoregressive model. An alternative direction for future research may be to avoid the causality issue altogether by introducing yet more complexity into strictly hierarchical priors.

\printbibliography

\clearpage

\begin{appendices}

\section{Rate-Distortion Curves}

\subsection{PSNR on Kodak}
\begin{figure}[ht]
\centering
\includegraphics[width=\linewidth]{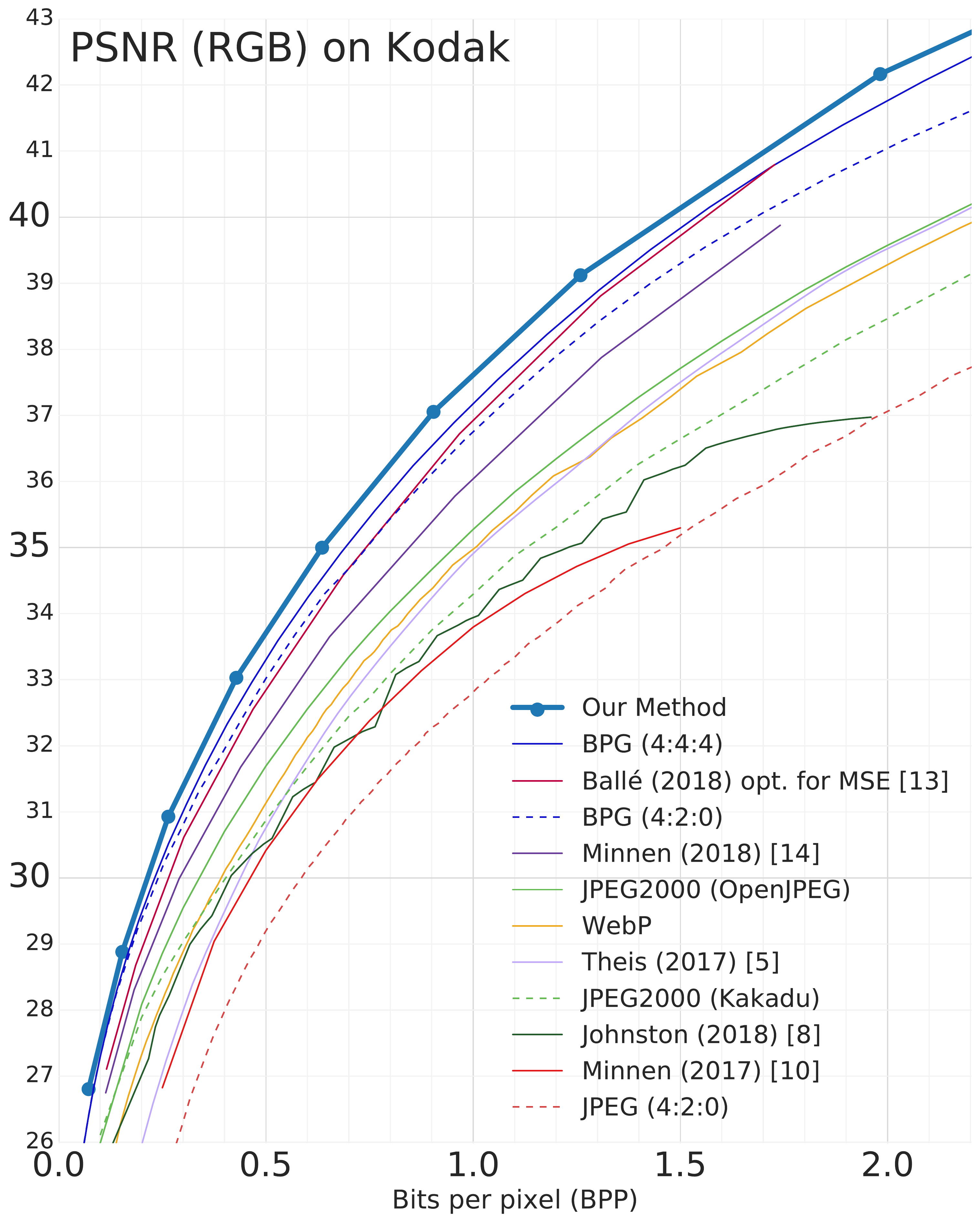}
\caption{When trained for MSE, our method provides better rate-distortion
  performance compared to all of the baseline methods when evaluated on the
  Kodak image set using PSNR. Each point on the RD curves is calculated by
  averaging over the PSNR and bit rate for the 24 Kodak images for a single Q
  value (for standard codecs) or $\lambda$ (for learned methods).}
\label{fig:kodak-psnr-rgb}
\end{figure}
\newpage

\subsection{MS-SSIM on Kodak}
\begin{figure}[ht]
\centering
\includegraphics[width=\linewidth]{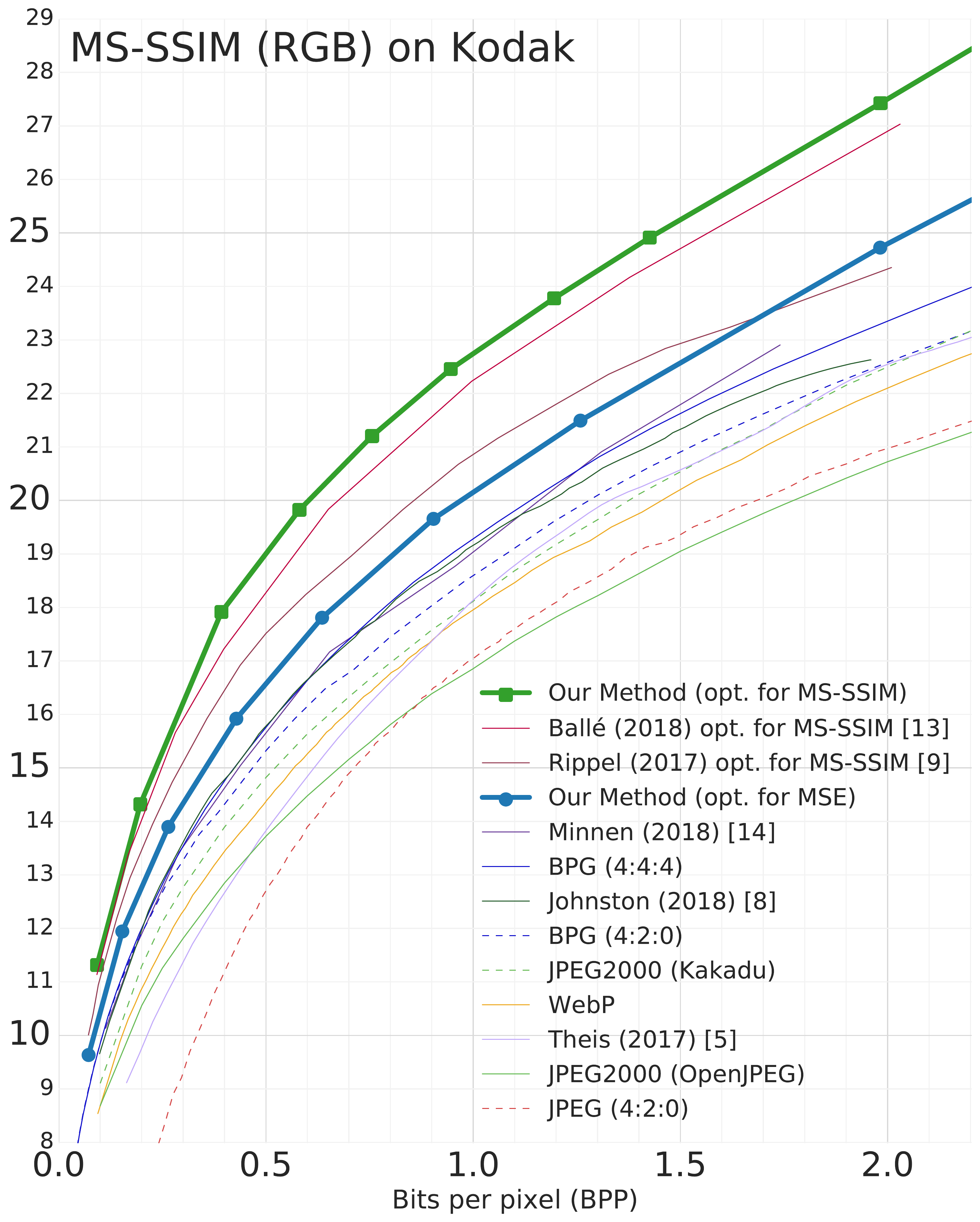}
\caption{When trained for MS-SSIM, our method provides better rate-distortion
  performance compared to all of the baseline methods when evaluated on the
  Kodak image set using MS-SSIM. Note that when our method is trained for MSE,
  the MS-SSIM score is still competitive. Specifically, it has higher MS-SSIM
  scores than any of the standard codecs and all learned methods that aren't
  optimized specifically for MS-SSIM. To improve readability, this graph shows
  MS-SSIM scores in dB using the formula: $\text{MS-SSIM}_{\text{dB}} = -10
  \log_{10}(1 - \text{MS-SSIM})$.}
\label{fig:kodak-msssim-rgb}
\end{figure}
\newpage

\subsection{SSIM on Kodak}
\begin{figure}[ht]
\centering
\includegraphics[width=\linewidth]{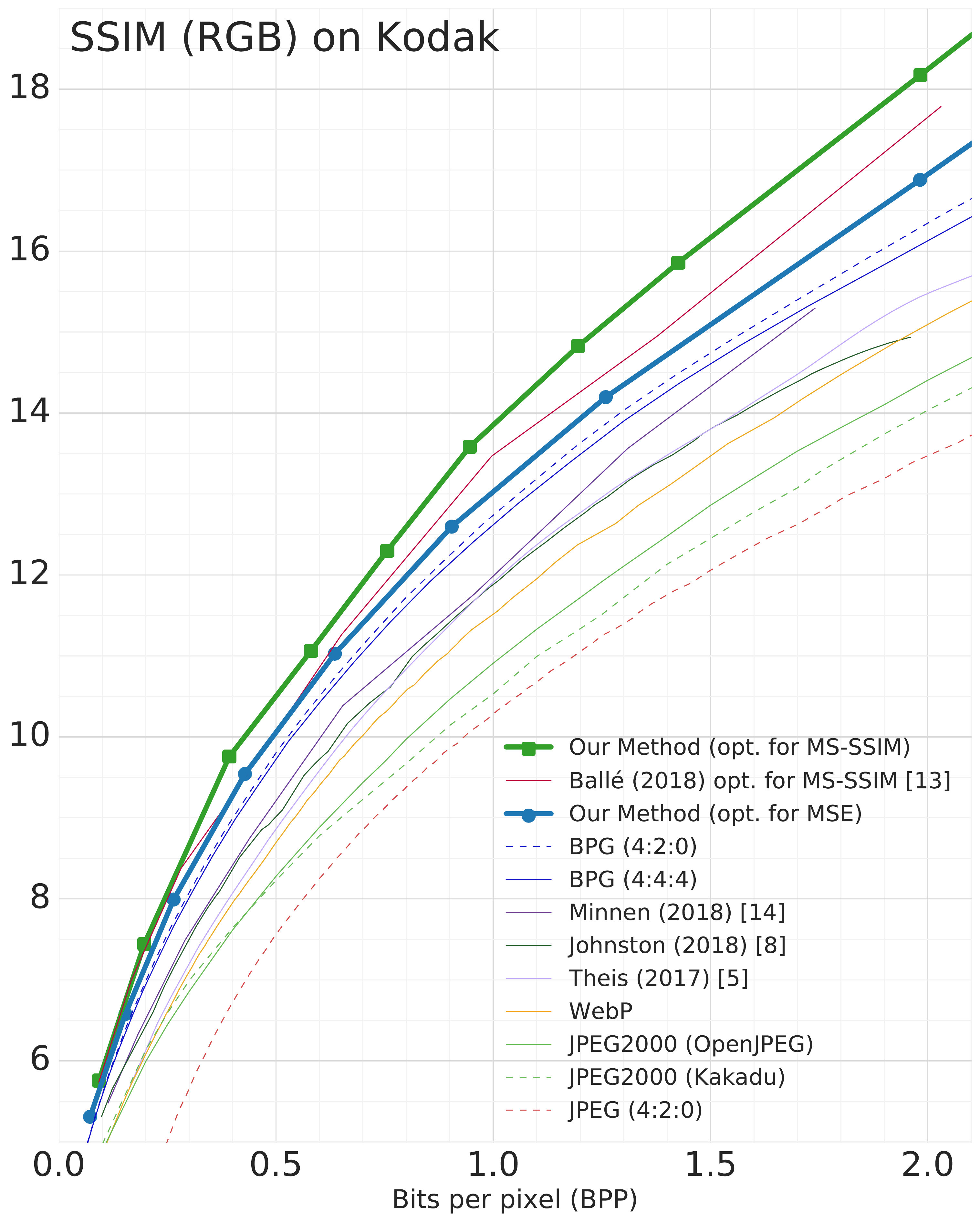}
\caption{Our method performs very well according to SSIM when optimized for
  MS-SSIM or MSE. Both versions outperform all of the baseline methods, except
  for one inversion above 1.1 bpp where the MSE-optimized version is worse
  than Ballé~\etal when that method is optimized for MS-SSIM. To improve
  readability, this graph shows SSIM scores in dB using the formula:
  $\text{SSIM}_{\text{dB}} = -10 \log_{10}(1 - \text{SSIM})$.}
\label{fig:kodak-ssim-rgb}
\end{figure}
\newpage

\subsection{PSNR on Tecnick}
\begin{figure}[ht]
\centering
\includegraphics[width=\linewidth]{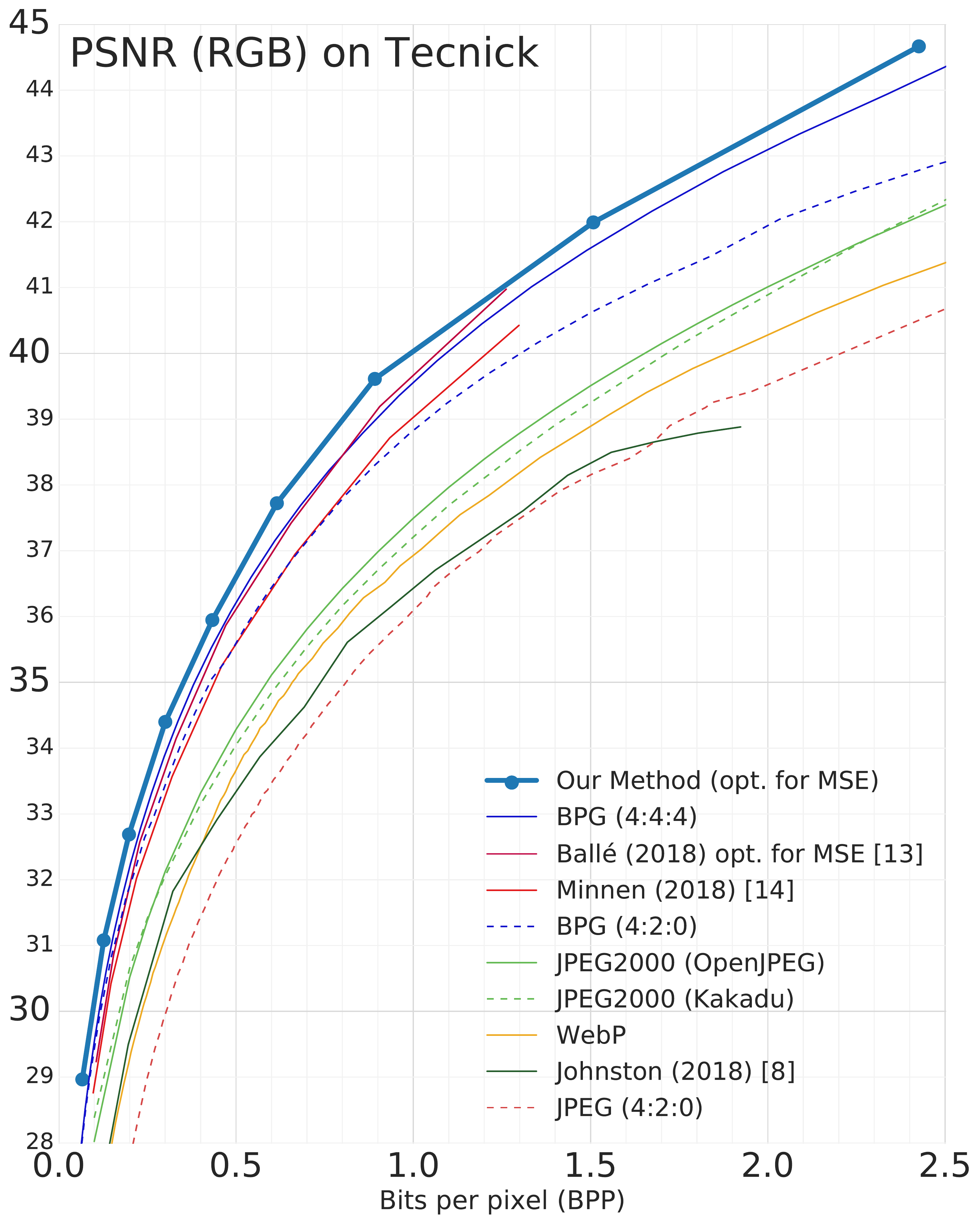}
\caption{When trained for MSE, our method provides better rate-distortion
  performance compared to all of the baseline methods when evaluated on the
  Tecnick image set using PSNR. Each point on the RD curves is calculated by
  averaging over the PSNR and bit rate values for the 100 Tecnick images for a
  single Q value (for standard codecs) or $\lambda$ (for learned methods). Each
  Tecnick image has resolution 1200$\times$1200.}
\label{fig:tecnick-psnr-rgb}
\end{figure}
\newpage

\section{Summary of Rate Savings}

\begin{figure}[ht]
\centering
\includegraphics[width=0.89\linewidth]{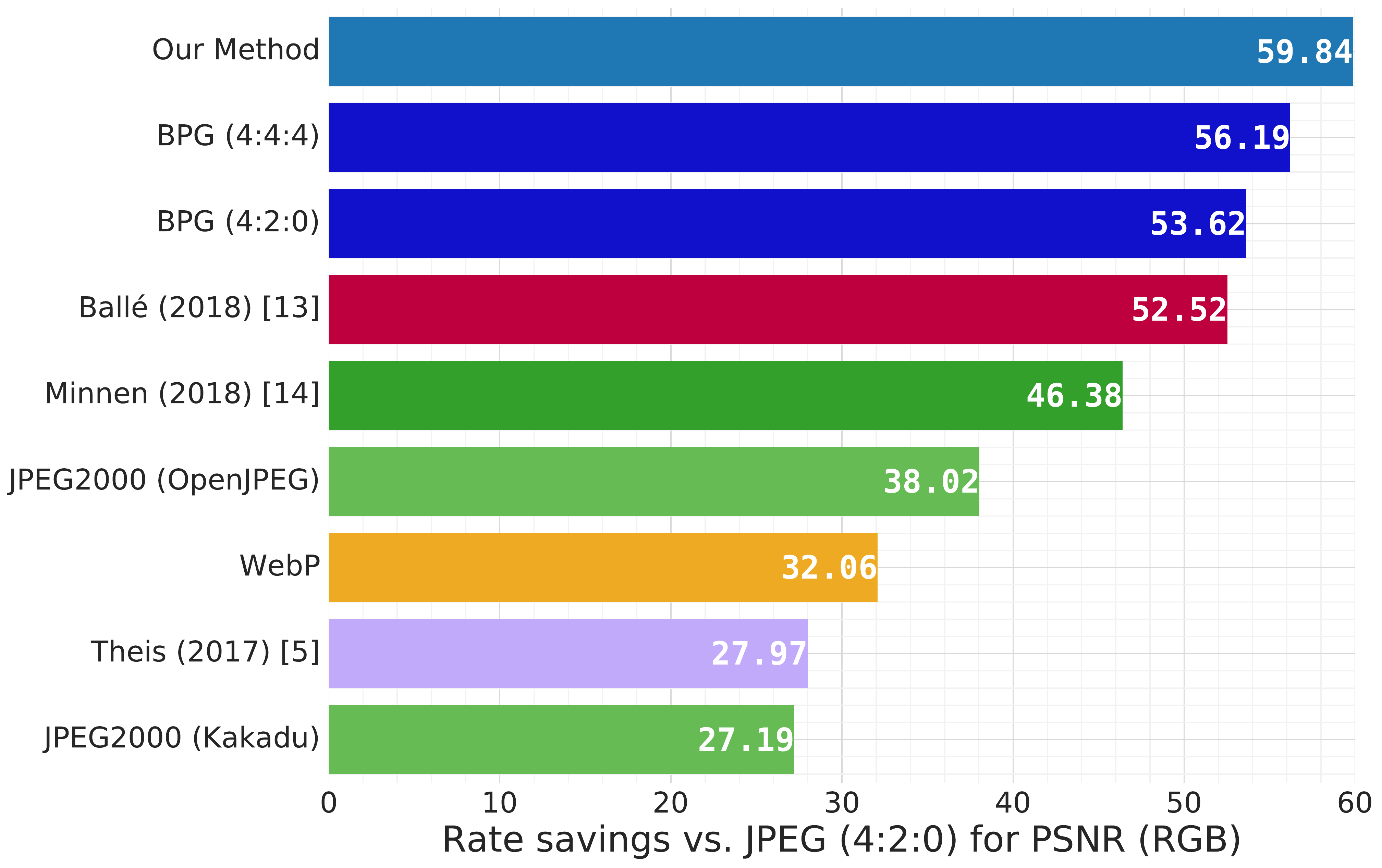}
\caption{This graph shows the average rate savings of each method compared to
  JPEG (4:2:0) using PSNR on the Kodak image set. Higher scores correspond to
  larger rate savings compared to JPEG as the baseline codec. These scores are
  calculated from the RD graph for PSNR on Kodak (see
  Figure~\ref{fig:kodak-psnr-rgb}) over the shared PSNR range of 27.1--39.9
  dB.}
\label{fig:bd-kodak-psnr-vs-jpeg}
\end{figure}

\begin{figure}[ht]
\centering
\includegraphics[width=0.89\linewidth]{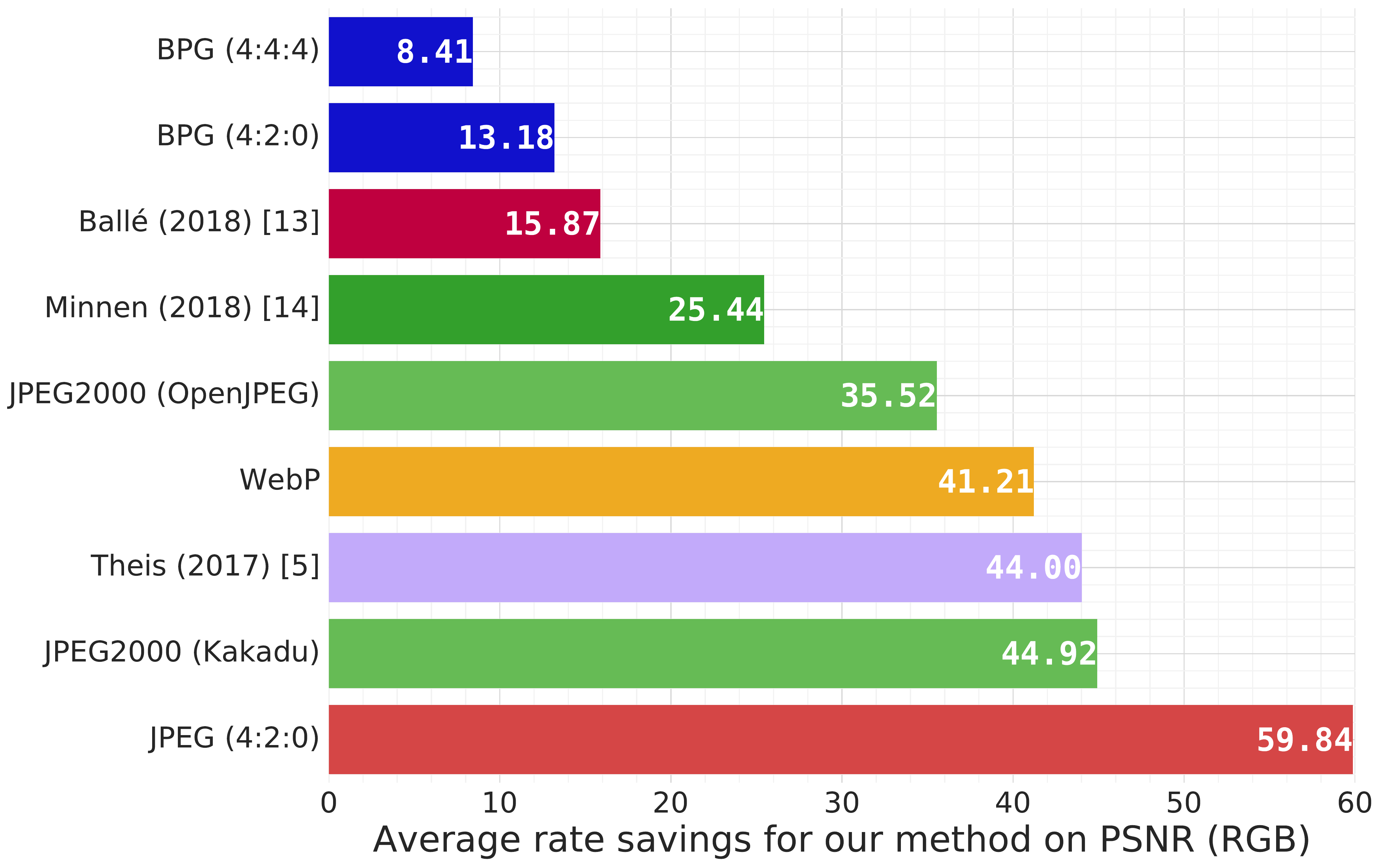}
\caption{This graph shows the average rate savings of our model compared to
  other codecs. Larger values imply a larger savings, \eg, on average, our
  method saves 8.41\% over BPG (4:4:4) and 35.52\% over JPEG2000 (based on the
  OpenJPEG implementation). These scores are calculated from the RD graph for
  PSNR on Kodak (see Figure~\ref{fig:kodak-psnr-rgb}) over the shared PSNR
  range of 27.1--39.9 dB.}
\label{fig:bd-kodak-psnr-vs-others}
\end{figure}
\newpage

\section{Example Images}

\subsection{Kodak 15}
\begin{figure}[ht]
\centering
\renewcommand{\tabcolsep}{2pt}
\begin{tabular}{ccc}  
  \includegraphics[width=0.49\linewidth]{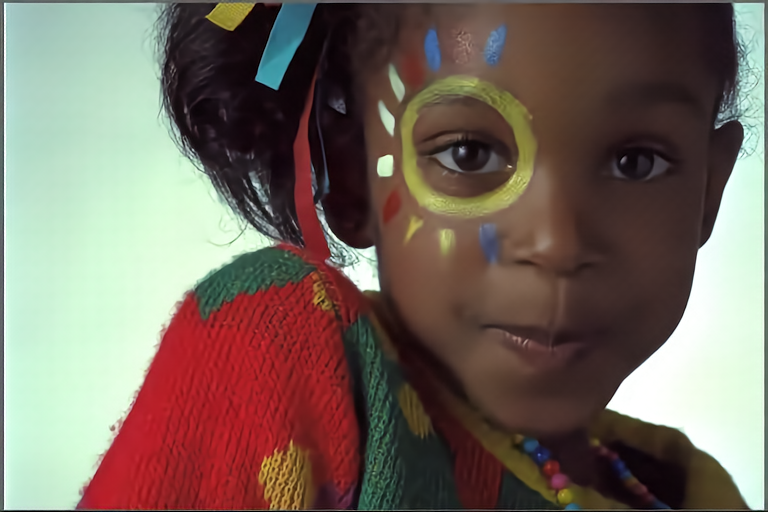}
  & \includegraphics[width=0.49\linewidth]{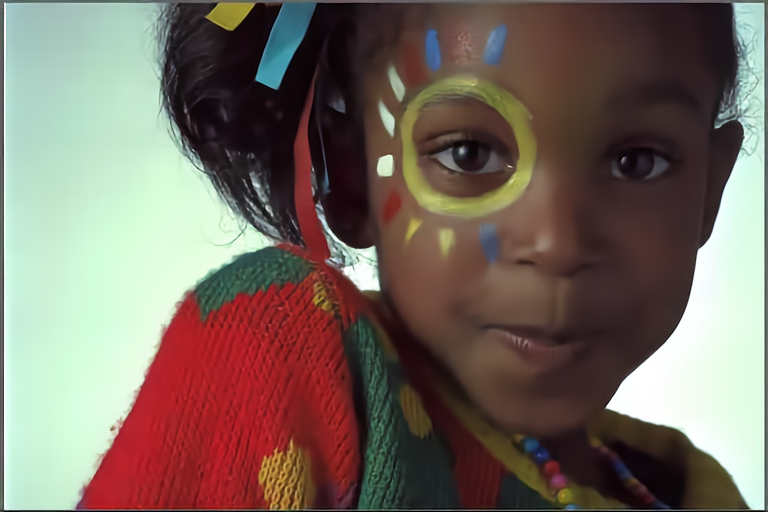}\\
  (a) Our Method (0.1552 bpp)
  & (b) Scale-only (0.1788 bpp)\\
  &&\\
  \includegraphics[width=0.49\linewidth]{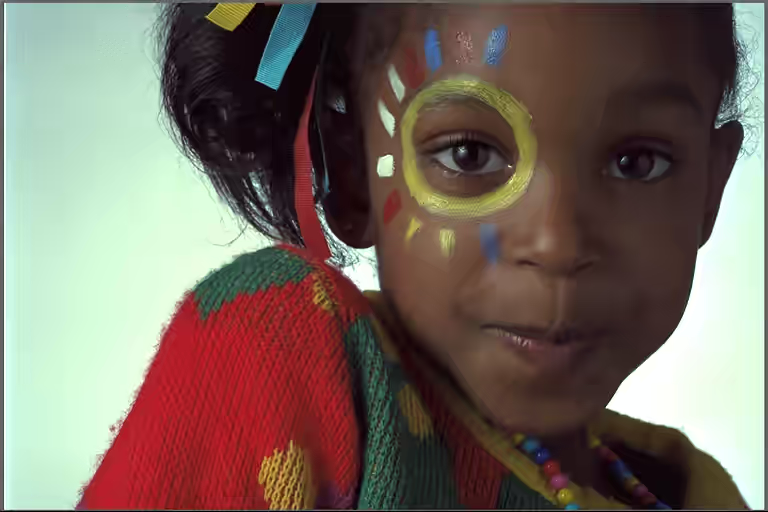}
  & \includegraphics[width=0.49\linewidth]{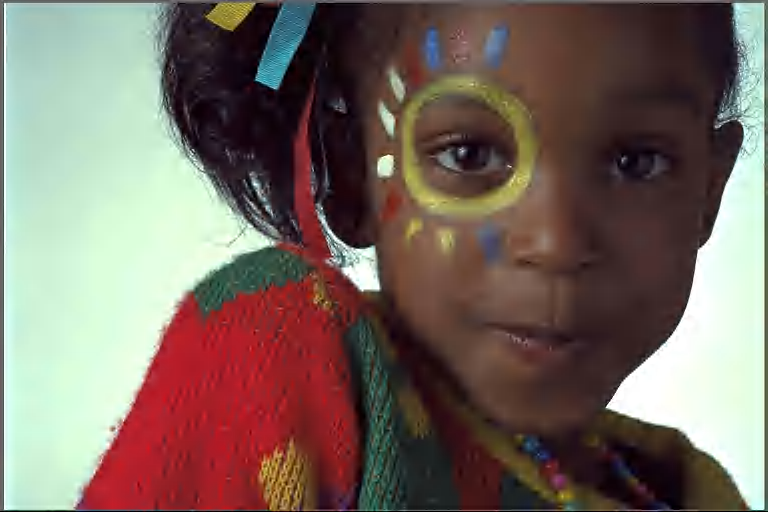}\\
  (c) BPG (0.1737 bpp)
  & (d) JPEG2000 (0.1557 bpp)\\
  &&\\
  \includegraphics[width=0.49\linewidth]{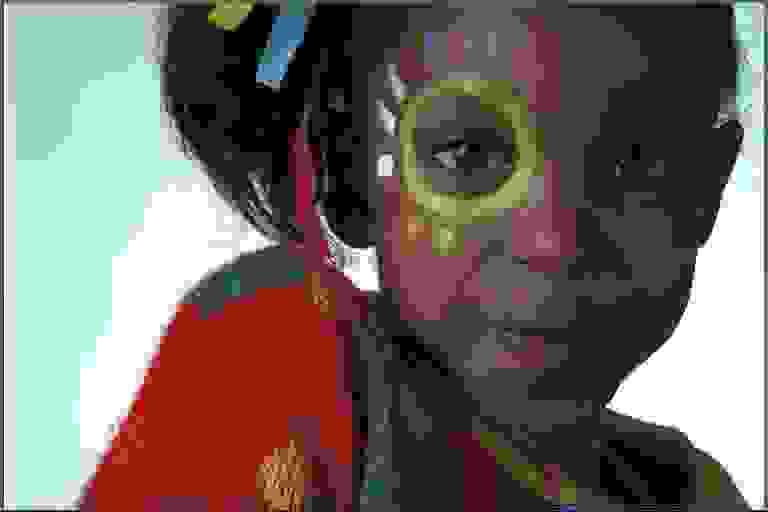}
  & \includegraphics[width=0.49\linewidth]{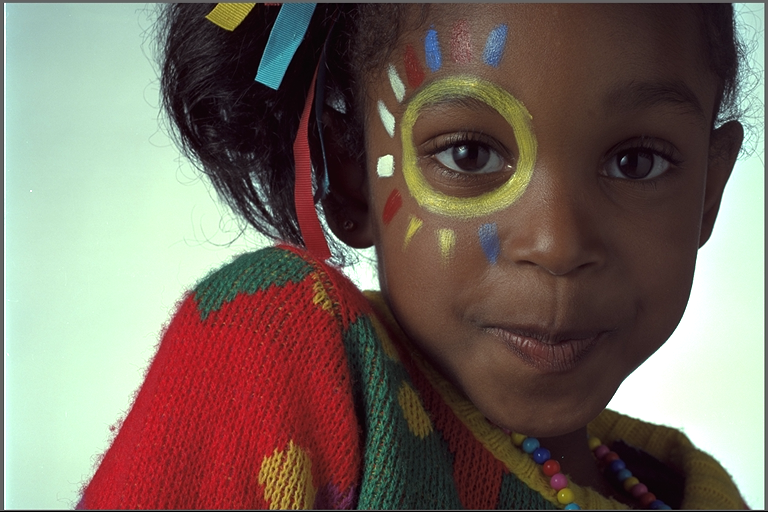}\\
  (e) JPEG (0.1634 bpp)
  & (f) Original
\end{tabular}  
  \caption{At similar bit rates, our method provides the highest visual
    quality on the Kodak 15 image. Note the geometric artifacts in BPG, \eg,
    on the girl's chin and cheek near the yellow face paint. JPEG2000 has
    severe artifacts and blurriness, while JPEG captures very little visual
    information at this bit rate. The reconstruction by the scale-only model
    is quite good, but it maintains less sharpness than our reconstruction
    (\eg, in the face paint and hair). Also note that both learning-based
    methods have a slight texture in the background, which is not in the
    original. We believe this arises due to the convolutional structure of the
    decoder network but is ultimately due to the use of mean squared error as
    the loss function, which is not very sensitive to subtle color shifts.}
  \label{fig:girl}
\end{figure}
\newpage

\subsection{Kodak 19}
\begin{figure}[ht]
\centering
\renewcommand{\tabcolsep}{2pt}
\begin{tabular}{ccc}  
  \includegraphics[width=0.32\linewidth]{figures/kodim19-nibbler-context-0_21493.png}
  & \includegraphics[width=0.32\linewidth]{figures/kodim19-nibbler-scale-0_22051.png}
  & \includegraphics[width=0.32\linewidth]{figures/kodim19-bpg-444-q36-0_23517.png}\\
  (a) Our Method (0.2149 bpp)
  & (b) Scale-only (0.2205 bpp)
  & (c) BPG (0.2352 bpp) \\
  & & \\
  \includegraphics[width=0.32\linewidth]{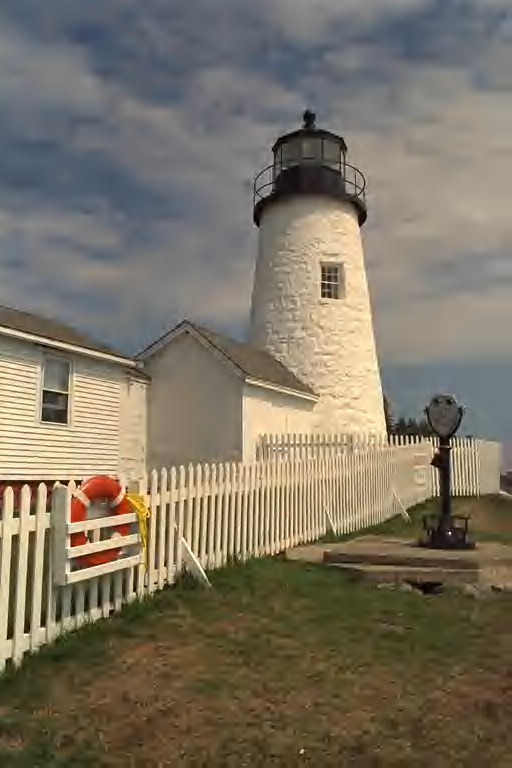}
  & \includegraphics[width=0.32\linewidth]{figures/kodim19-jpg-420-q006-0_23094.jpg}
  & \includegraphics[width=0.32\linewidth]{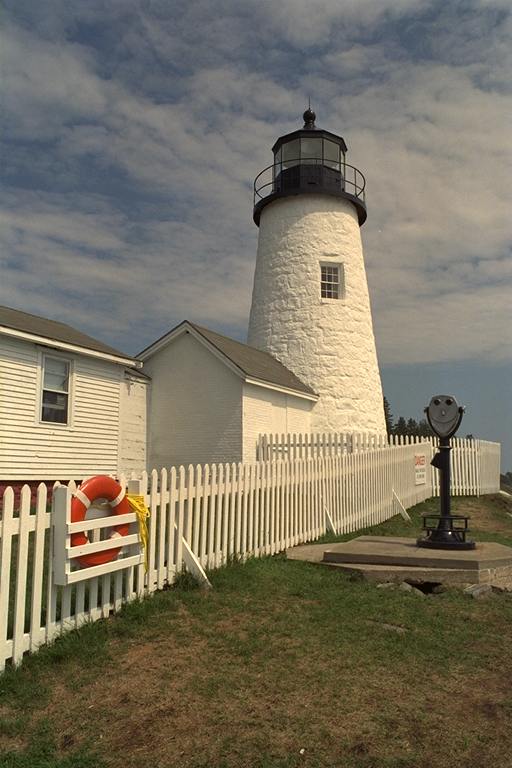}\\
  (d) JPEG2000 (0.2152 bpp)
  & (e) JPEG (0.2309 bpp)
  & (f) Original
\end{tabular}
  \caption{At similar bit rates, our method provides the highest visual
    quality on the Kodak 19 image. Note the aliasing in the fence in the
    scale-only version as well as a slight global color cast and blurriness in
    the yellow rope. BPG shows more ``classical'' compression artifacts, \eg,
    ringing around the top of the lighthouse and the roof of the middle
    building. BPG also introduces a few geometric artifacts in the sky, though
    it does preserve extra detail in the sky and fence (albeit with 9.4\% more
    bits that our encoding). JPEG shows severe blocking artifacts at this bit
    rate, and JPEG2000 includes many artifacts in the sky and near object
    boundaries.}
  \label{fig:lighthouse-appendix}
\end{figure}
\newpage

\subsection{Kodak 20}
\begin{figure}[ht]
\centering
\renewcommand{\tabcolsep}{2pt}
\begin{tabular}{ccc}  
  \includegraphics[width=0.49\linewidth]{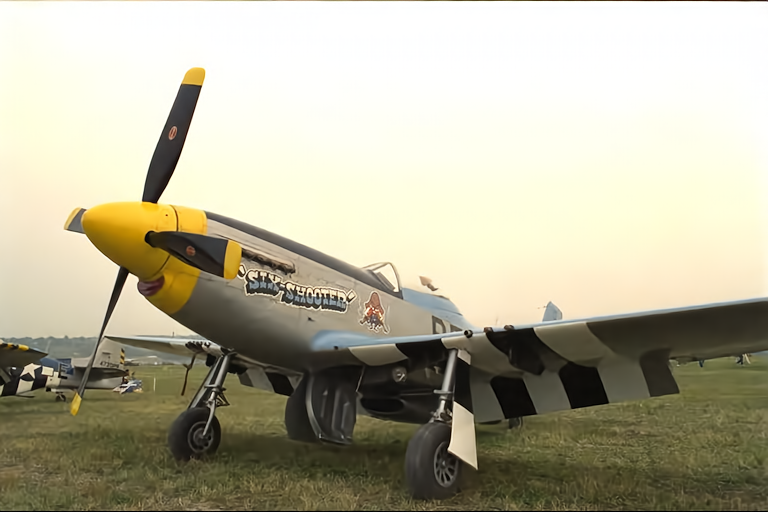}
  & \includegraphics[width=0.49\linewidth]{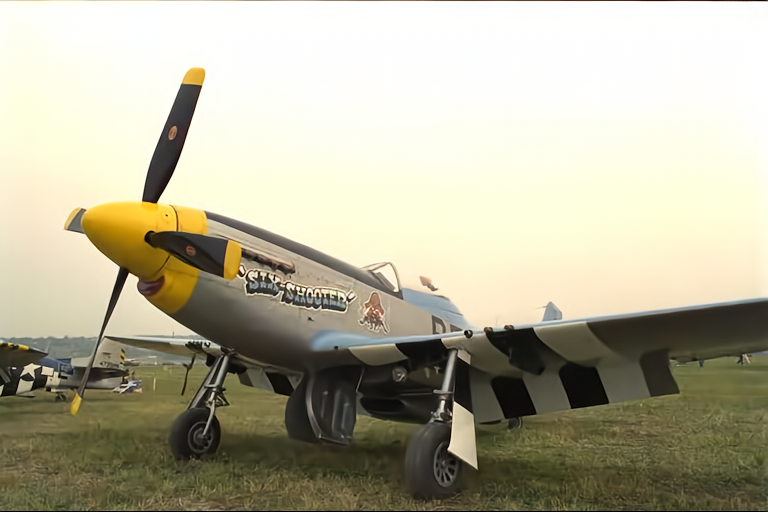}\\
  (a) Our Method (0.2464 bpp)
  & (b) Scale-only (0.2760 bpp)\\
  &&\\
  \includegraphics[width=0.49\linewidth]{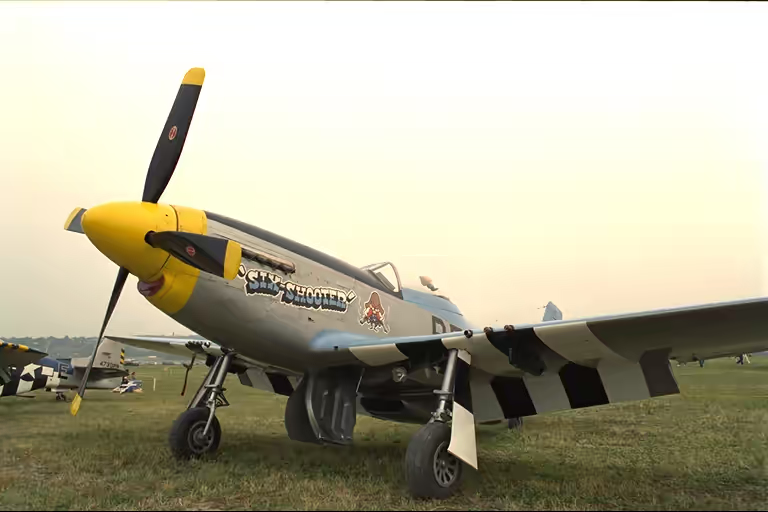}
  & \includegraphics[width=0.49\linewidth]{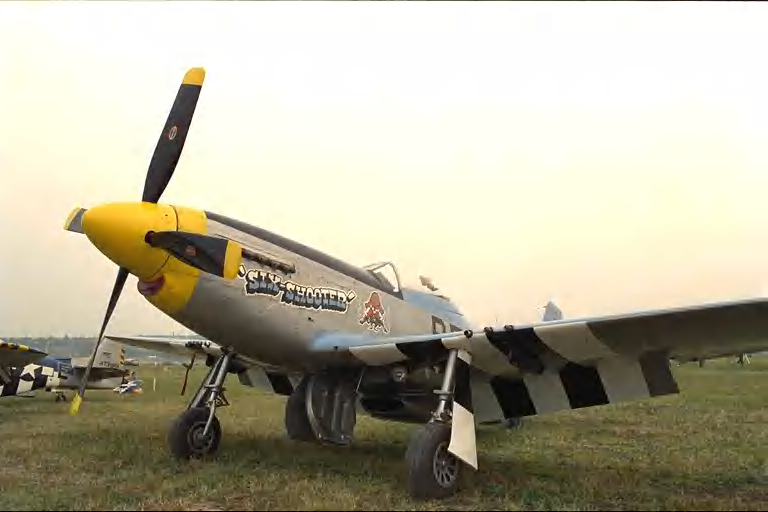}\\
  (c) BPG (0.2565 bpp)
  & (d) JPEG2000 (0.2499 bpp)\\
  &&\\
  \includegraphics[width=0.49\linewidth]{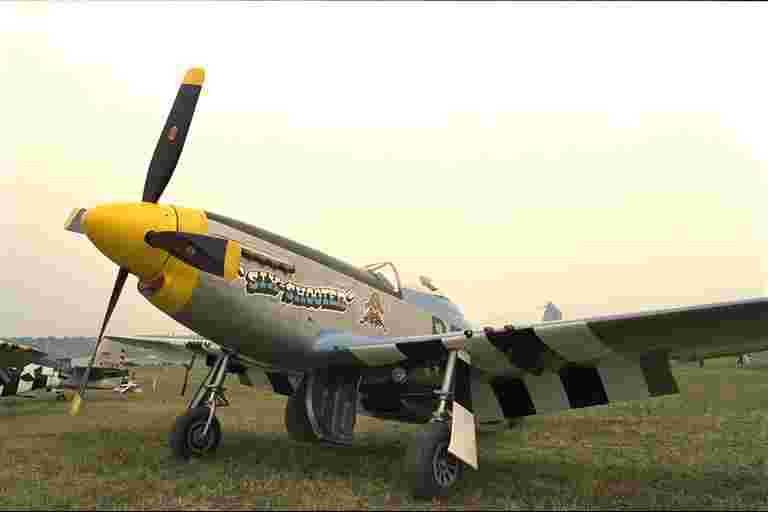}
  & \includegraphics[width=0.49\linewidth]{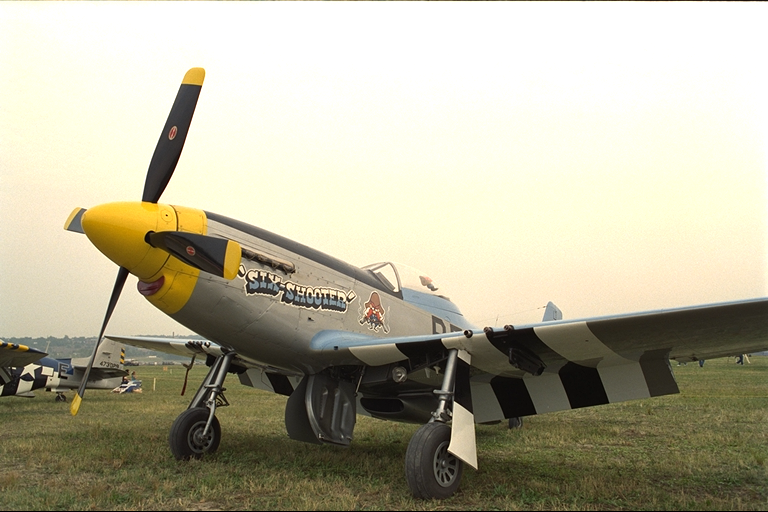}\\
  (e) JPEG (0.2521 bpp)
  & (f) Original
\end{tabular}  
  \caption{At similar bit rates, our method provides the highest visual
    quality on the Kodak 20 image. Note the artifacts in the sky in the BPG
    reconstruction and severe artifacts for JPEG and JPEG2000. The scale-only
    reconstruction is much closer visually, but our method has slightly more
    sharpness, \eg, in the red circles on the propeller and in the Yosemite
    Sam nose art, despite the fact that our method uses more than 10\% fewer
    bits.}
  \label{fig:plane}
\end{figure}
\newpage

\subsection{Kodak 23}
\begin{figure}[ht]
\centering
\renewcommand{\tabcolsep}{2pt}
\begin{tabular}{ccc}  
  \includegraphics[width=0.49\linewidth]{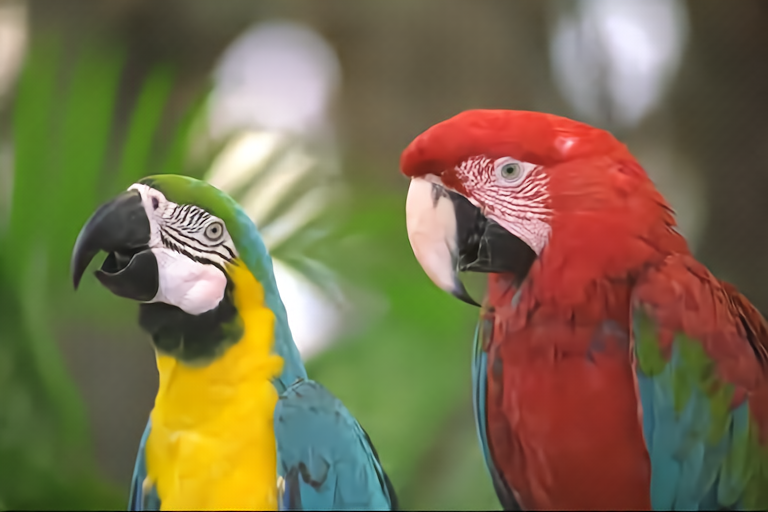}
  & \includegraphics[width=0.49\linewidth]{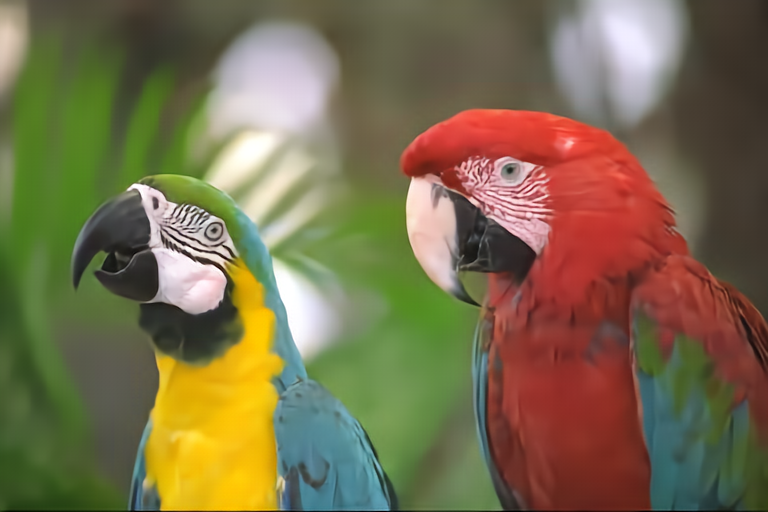}\\
  (a) Our Method (0.1227 bpp)
  & (b) Scale-only (0.1378 bpp)\\
  &&\\
  \includegraphics[width=0.49\linewidth]{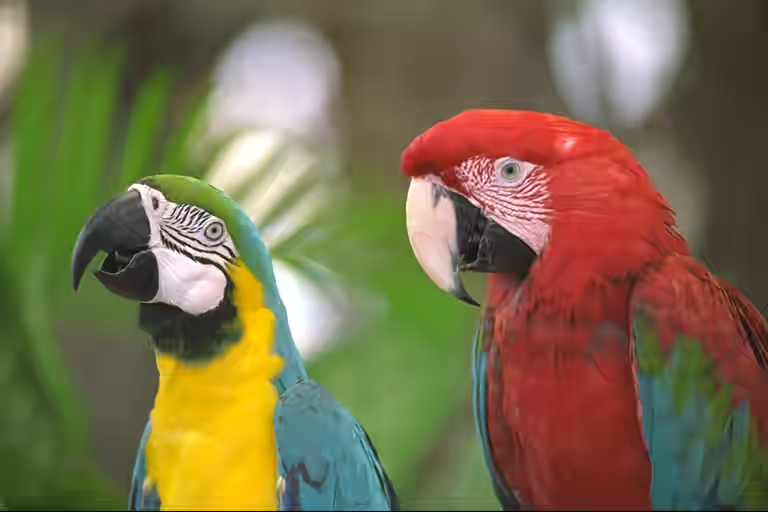}
  &   \includegraphics[width=0.49\linewidth]{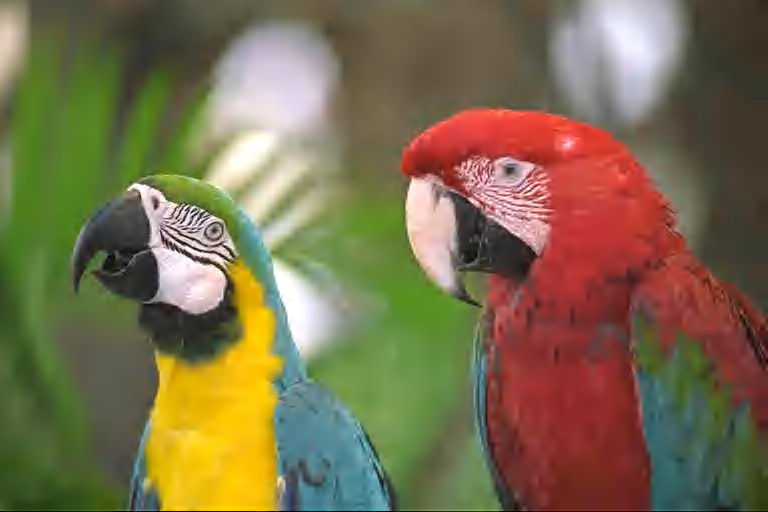}\\
  (c) BPG (0.1293 bpp)
  & (d) JPEG2000 (0.1241 bpp)\\
  &&\\
  \includegraphics[width=0.49\linewidth]{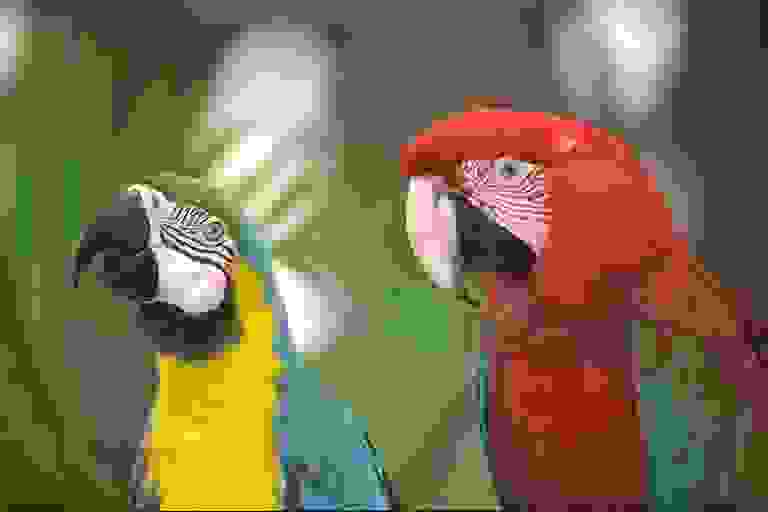}
  & \includegraphics[width=0.49\linewidth]{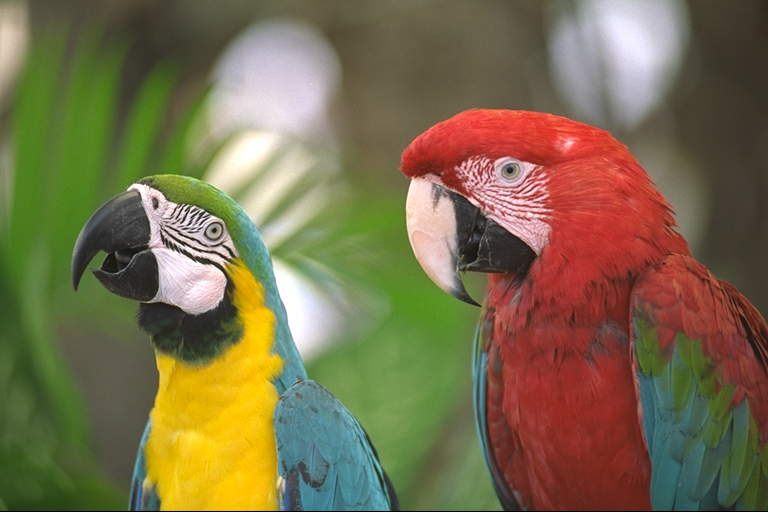}\\
  (e) JPEG (0.1576 bpp)
  & (f) Original
\end{tabular}  
  \caption{At similar bit rates, our method provides the highest visual
    quality on the Kodak 23 image. Note that the BPG reconstruction has some
    ringing and geometric artifacts (\eg, at the top of the red parrot's
    head), while JPEG2000 has many artifacts around object boundaries. The
    scale-only method has similar visual quality but is blurrier than our
    reconstruction, \eg, in the red parrot's eye and face.}
  \label{fig:parrots}
\end{figure}
\newpage

\section{Architecture Comparison}

\begin{figure}[ht]
\centering
\includegraphics[width=0.85\linewidth]{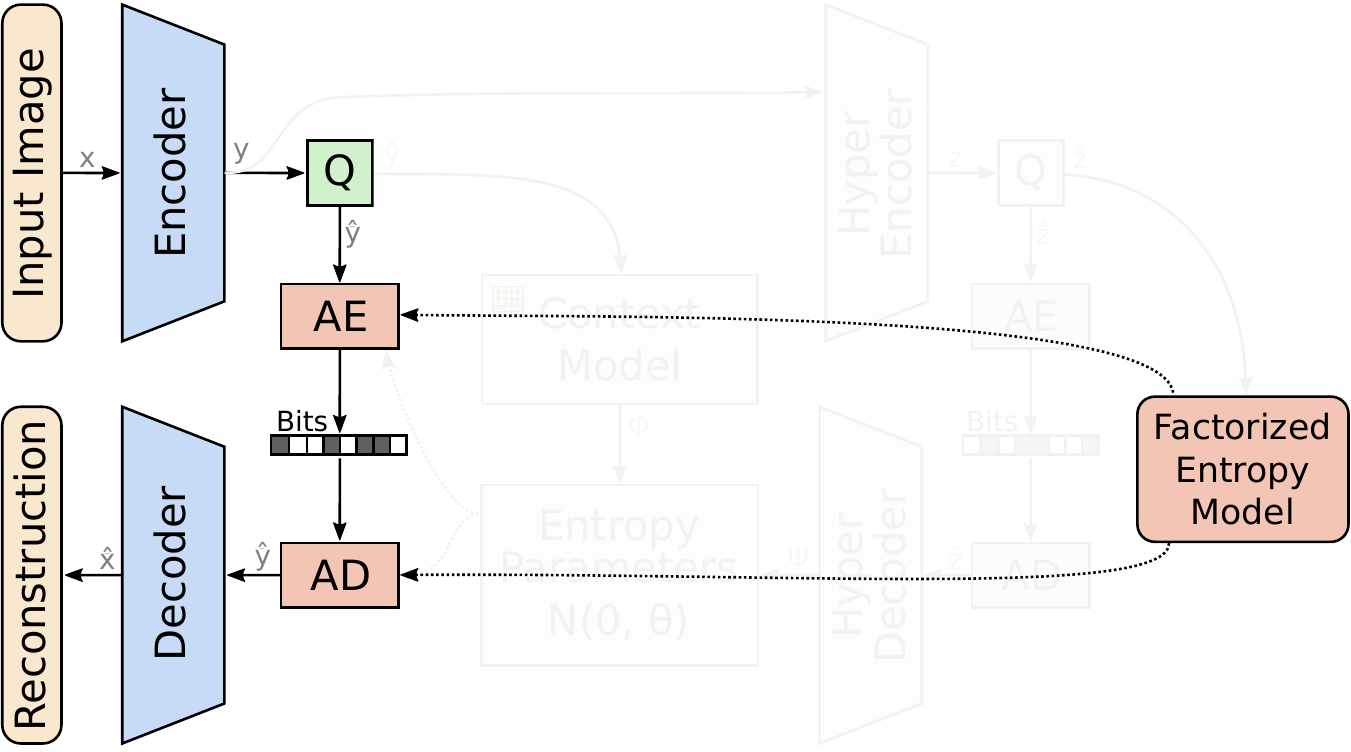}
\caption{\textbf{Fully-Factorized Model} This model learns a fixed entropy
  model that is shared between the encoder and decoder systems. It is designed
  based on the assumption that all latents are i.i.d and ideally set to be a
  multinomial. In practice, it must be relaxed to ensure
  differentiability. The compression model is primarily an autoenocder
  composed of convolutional layers and nonlinear activations, where the main
  complication is due to ensuring that the entropy model is differentiable and
  that useful gradients flow through the quantized latent representation.}
\label{fig:arch-factorized}
\end{figure}

\begin{figure}[ht]
\centering
\includegraphics[width=0.85\linewidth]{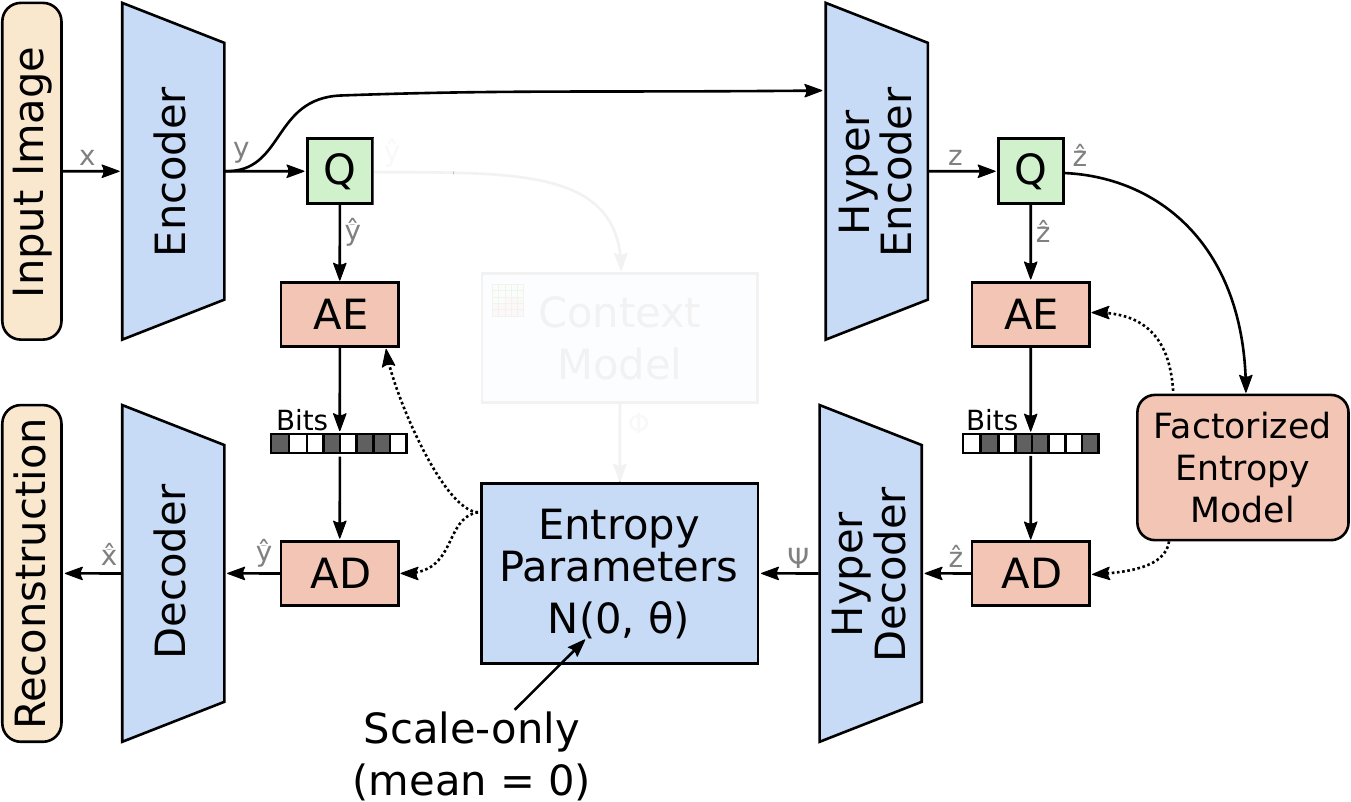}
\caption{\textbf{Scale-only Hyperprior} This model uses a conditional Gaussian
  scale mixture (GSM) as the entropy model. The GSM is conditioned on a
  learned hyperprior, which is a (hyper-)latent representation formed by
  transforming the latents using the \textit{Hyper-Encoder}. The
  \textit{Hyper-Decoder} can then decode the hyperprior to create the scale
  parameters for the GSM. The main advantage of this model is that the entropy
  model is image-dependent and can be adapted for each individual code. The
  downside is that the compressed hyperprior must be transmitted with the
  compressed latents, which increases the total file size.}
\label{fig:arch-scale}
\end{figure}

\begin{figure}[ht]
\centering
\includegraphics[width=0.85\linewidth]{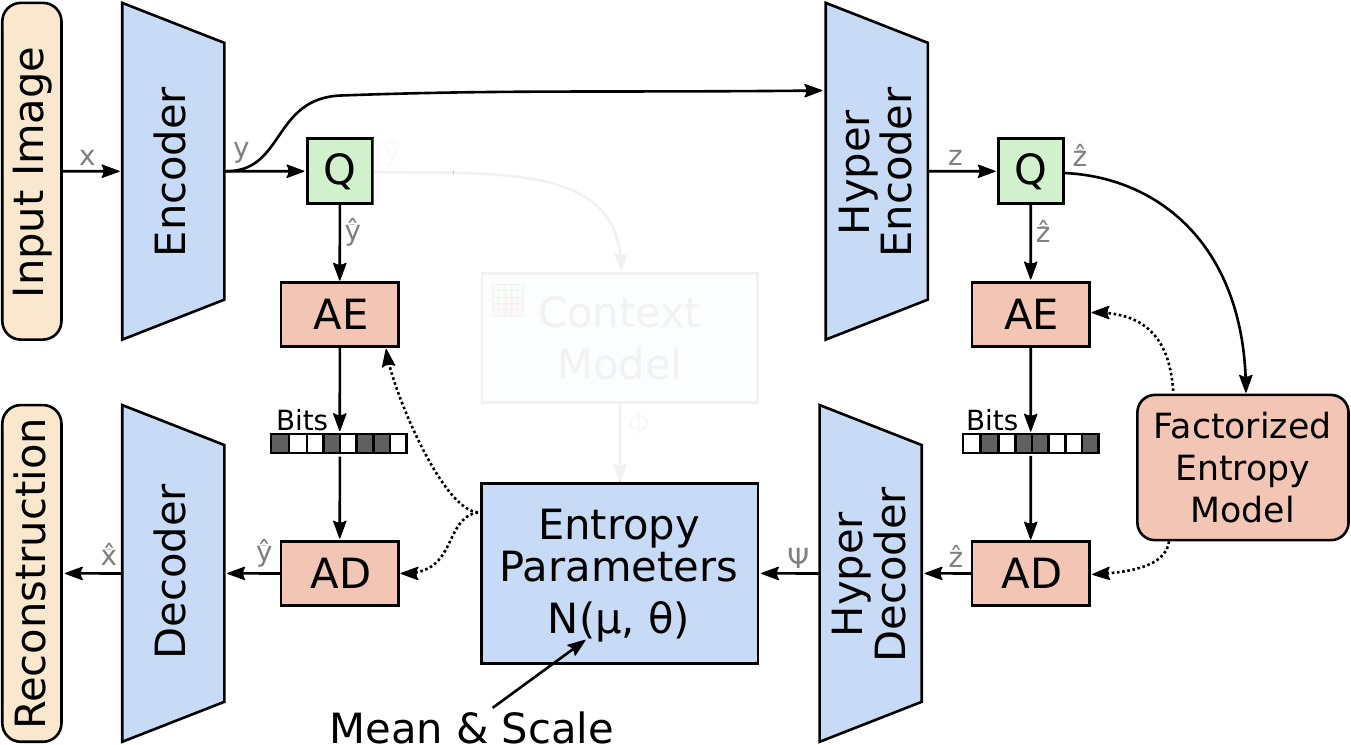}
\caption{\textbf{Mean \& Scale Hyperprior} This model variant is a simple
  extension of the scale-only hyperprior model shown in
  Figure~\ref{fig:arch-scale} in which the GSM is replaced with a Gaussian
  mixture model (GMM). The \textit{Hyper-Decoder} is therefore responsible for
  transforming the hyperprior into both the mean and scale parameters of the
  Gaussians.}
\label{fig:arch-mean}
\end{figure}

\begin{figure}[ht]
\centering
\includegraphics[width=0.85\linewidth]{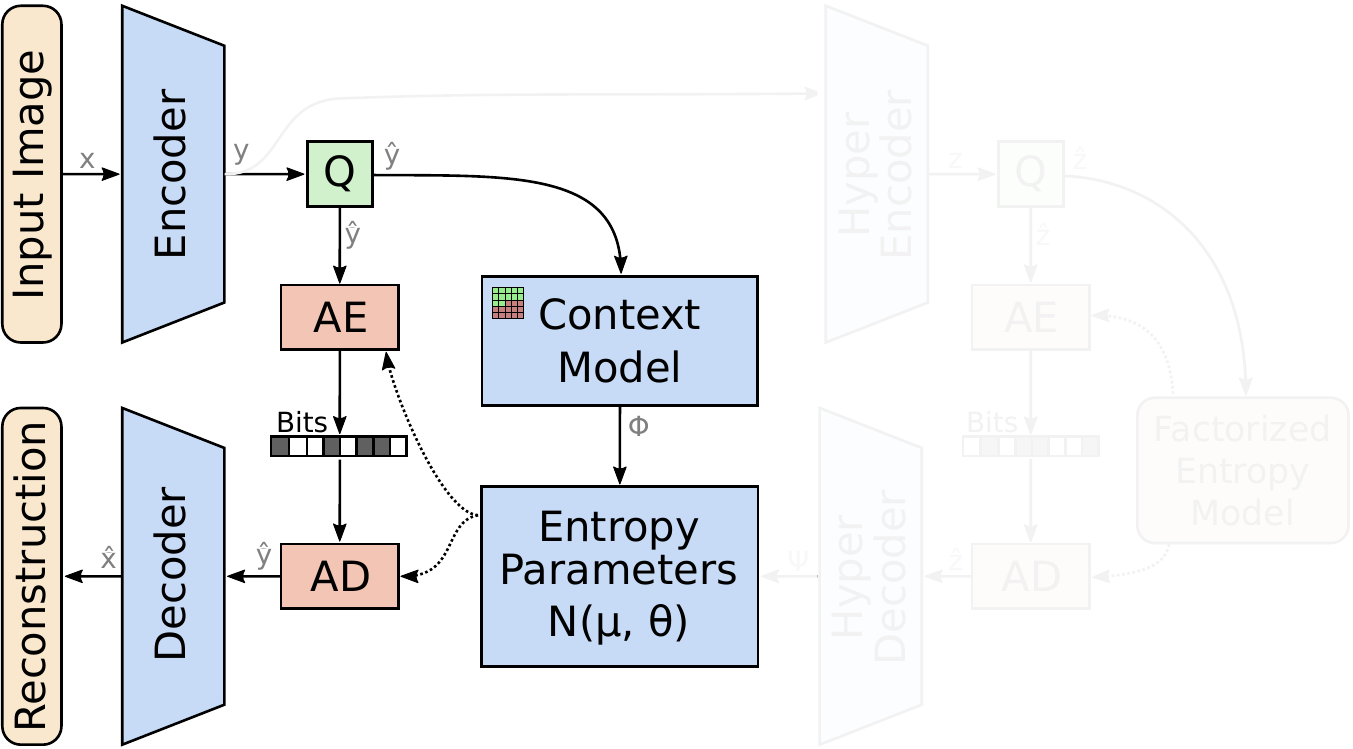}
\caption{\textbf{Context-only Model} This model does not use a hyperprior and
  instead relies only on an autoregressive process to predict the parameters
  of the GMM-based entropy model. The benefit of this approach is that no
  additional bits are added to the bit-stream. The downside is that the
  autoregressive model can only access codes in its causual context since the
  decoder runs serially in raster-scan order.}
\label{fig:arch-cxtonly}
\end{figure}

\begin{figure}[ht]
\centering
\includegraphics[width=0.85\linewidth]{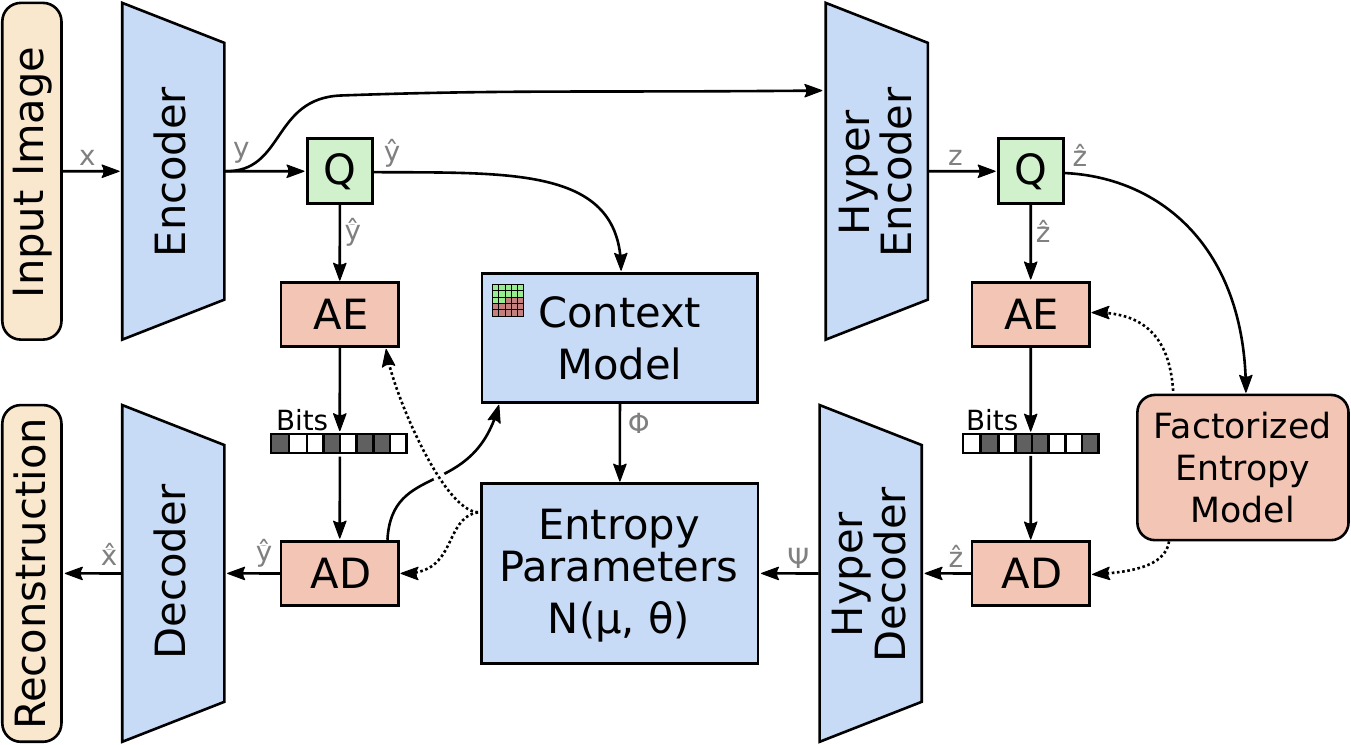}
\caption{\textbf{Context + Hyperprior} By combining the mean \& scale
  hyperprior with an autoregressive model, we form the full model presented in
  the paper. Our evaluation shows that the autoregressive model and the
  hyperprior are complementary, and that joint optimization leads to a
  compression model with better rate-distortion performance than either
  approach on its own.}
\label{fig:arch-context-and-hyperprior}
\end{figure}

\end{appendices}

\end{document}